%% file: main.tex
\documentclass[11pt]{article}

\usepackage{acl}

\usepackage{times}
\usepackage{latexsym}
\usepackage[T1]{fontenc}

\usepackage[utf8]{inputenc}

\usepackage{microtype}

\usepackage{inconsolata}

\usepackage{amsmath}
\usepackage{amssymb}
\usepackage{enumitem}

\usepackage{graphicx}
\usepackage{subcaption}  %

\usepackage{algorithm}
\usepackage{algorithmic}
\algsetup{linenosize=\small}  %

\usepackage{longtable}
\usepackage{booktabs}
\usepackage{tabularx}
\usepackage{array}
\usepackage{colortbl}
\usepackage{ltxtable}
\usepackage{comment}
\usepackage{multirow}
\usepackage{diagbox}
\usepackage{enumitem}
\usepackage{ragged2e}

\newcolumntype{Y}{>{\RaggedRight\arraybackslash}X}
\newcolumntype{L}[1]{>{\RaggedRight\arraybackslash}p{#1}}
\usepackage{xcolor}

\usepackage{filecontents}
\usepackage{hyperref}

\usepackage{graphicx}
\usepackage{subcaption}

\usepackage{tabularx}     
\usepackage{booktabs}      
\usepackage{enumitem}      

\usepackage{tcolorbox}
\tcbuselibrary{skins, breakable}

\newtcolorbox{systemcard}[1]{
    enhanced, breakable,
    colback=gray!2!white,         
    colframe=gray!60!black,       
    colbacktitle=gray!10!white,   
    coltitle=black,               
    fonttitle=\bfseries\sffamily, 
    title={#1},                   
    boxrule=0.5pt,                
    titlerule=0.5pt,              
    arc=3pt,
    left=8pt, right=8pt, top=6pt, bottom=6pt, 
    fontupper=\small,             
    before skip=6pt, after skip=6pt 
}

\usepackage{listings}
\definecolor{backcolour}{rgb}{0.97,0.97,0.97}
\definecolor{stringcolor}{rgb}{0.13,0.55,0.13} 
\definecolor{keycolor}{rgb}{0.1,0.1,0.6}       

\lstdefinestyle{jsonstyle}{
    backgroundcolor=\color{backcolour},   
    stringstyle=\color{stringcolor},
    keywordstyle=\color{keycolor},
    basicstyle=\ttfamily\scriptsize, 
    breakatwhitespace=false,         
    breaklines=true,      
    captionpos=b,                    
    keepspaces=true,                 
    showspaces=false,                
    showstringspaces=false,
    frame=single,                    
    rulecolor=\color{lightgray}
}

\newif\ifcomments
\commentstrue  %

\ifcomments
  \newcommand{\ZH}[1]{\textcolor[HTML]{228B22}{#1}} 
\else
  \newcommand{\ZH}[1]{}
\fi

\usepackage{color}

\newif\ifcr
\crtrue
\usepackage[normalem]{ulem}
\usepackage{makecell}
\ifcr
  
  \DeclareRobustCommand{\CD}[1]{\textcolor[HTML]{909090}{\sout{#1}}}
  \DeclareRobustCommand{\CN}[2]{\textcolor[HTML]{E07000}{\,{\footnotesize[#1: #2]}}}
  \newenvironment{CAblock}{\par\color[HTML]{C00000}}{\par}
\else
  
  \DeclareRobustCommand{\CD}[1]{}
  \DeclareRobustCommand{\CN}[2]{}
  
\fi

\title{SocialReasonBench: A Video-QA Benchmark for Social Reasoning \\ with Counterfactual Narrative Videos}

\author{
  \textbf{Zheyu Huang}\textsuperscript{1},
  \textbf{Zijing Shi}\textsuperscript{1},
  \textbf{Haozhe Luo}\textsuperscript{2},
  \textbf{Huadong Tang}\textsuperscript{1}, \\[2pt]
  \textbf{Mingyu Liu}\textsuperscript{3},
  \textbf{Meng Fang}\textsuperscript{4},
  \textbf{Ling Chen}\textsuperscript{1} \\[5pt]
  \textsuperscript{1}AAII, University of Technology Sydney \quad
  \textsuperscript{2}Northeastern University \\
  \textsuperscript{3}Fudan University \quad
  \textsuperscript{4}University of Liverpool \\[3pt]
  \texttt{\{Zheyu.Huang, Huadong.Tang\}@student.uts.edu.au} \\[1pt]
  \texttt{\{Zijing.Shi-1, Ling.Chen\}@uts.edu.au} \quad
  \texttt{202312249@stu.neu.edu.cn} \\
  \texttt{25213070025@m.fudan.edu.cn} \quad\texttt{Meng.Fang@liverpool.ac.uk}
}

\begin{document}

\maketitle

\begin{abstract}
Recent advances in Large Multimodal Models (LMMs) have greatly improved video understanding, yet their ability to reason about human-centered social situations remains limited. Existing benchmarks typically rely on videos with a single observed trajectory, making it difficult to determine whether models truly understand social dynamics or merely exploit recurring narrative patterns. We introduce SocialReasonBench, a video multiple-choice QA benchmark for evaluating socially grounded reasoning in scenarios derived from interactive narratives. Built from gameplay videos of \textit{Detroit: Become Human}, the benchmark leverages branching storylines where player decisions lead to alternative social outcomes that can be checked against the game's own script, flowchart, and recorded branches. We develop a multi-agent curation pipeline that localizes socially meaningful clips, grounds answer labels in game-state signals, and generates theory-guided questions with diagnostic distractors. SocialReasonBench covers seven reasoning dimensions, including intent recognition, emotional empathy, moral dilemma, counterfactual reasoning, and causal antecedent. Experiments on contemporary LMMs show that models perform reasonably well on basic social understanding but struggle with counterfactual and causal reasoning. Further ablation and diagnostic error analyses reveal that models often depend on incomplete modality cues and fall into reasoning traps such as visual shortcuts, highlighting a gap between observable event recognition and deeper reasoning over latent social states.

\end{abstract}

\input{chapters/introduction}

\input{chapters/related_works}

\input{chapters/methodology}

\input{chapters/experiments}

\input{chapters/results}
\input{chapters/conclusion}
\input{chapters/limitation}

\bibliography{bibliography}

\clearpage
\appendix
\input{chapters/appendix}

\end{document}

%% file: chapters/introduction.tex
\begin{figure}[t]
    \centering
     \includegraphics[width=\linewidth]{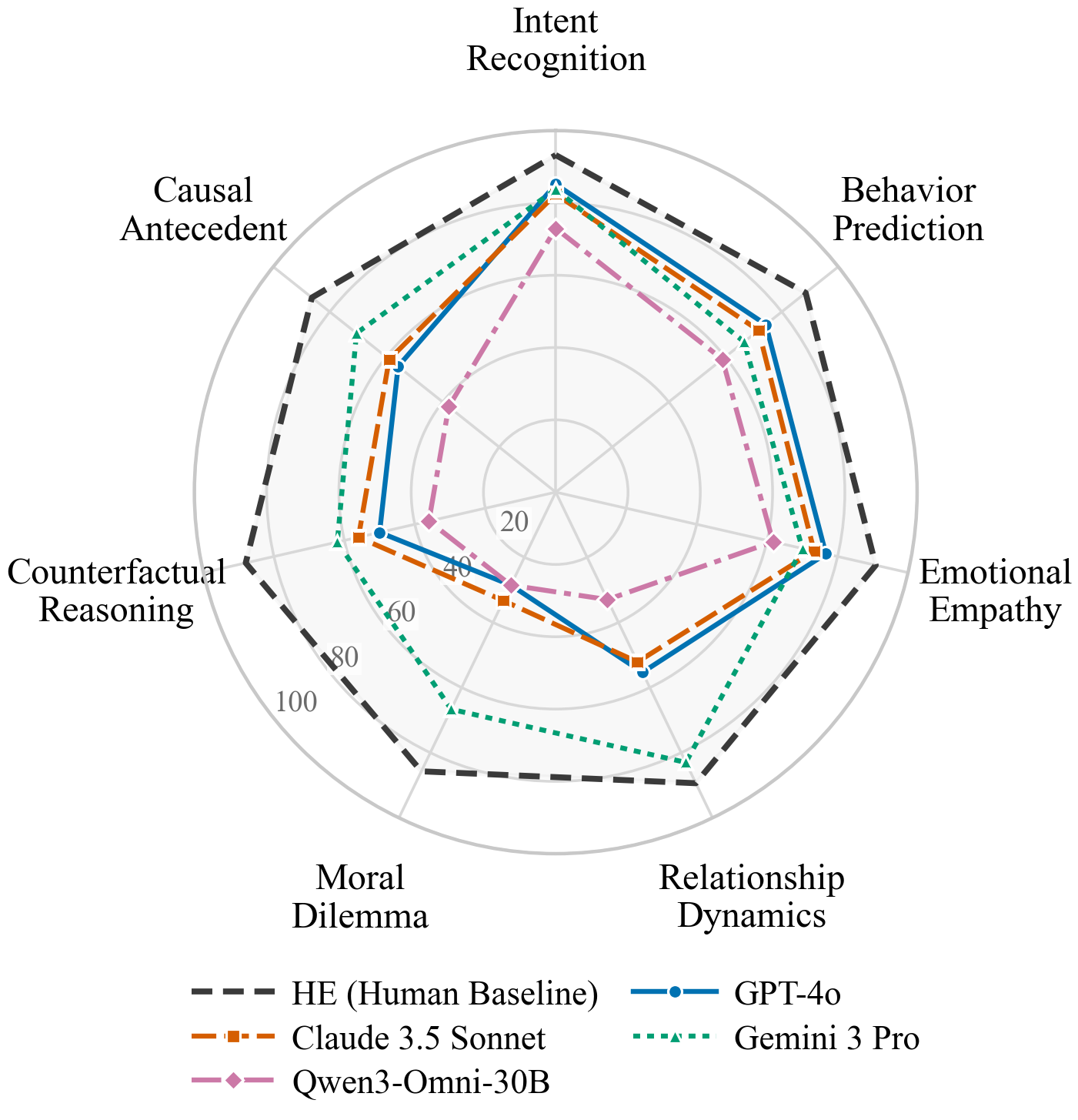}
    \vspace{-2mm}
    \caption{Model performance across SocialReasonBench dimensions.}
    \vspace{-6mm}
    \label{fig:radar_chart}
\end{figure}

\begin{figure*}[t]
  \centering
      \includegraphics[width=0.98\linewidth]{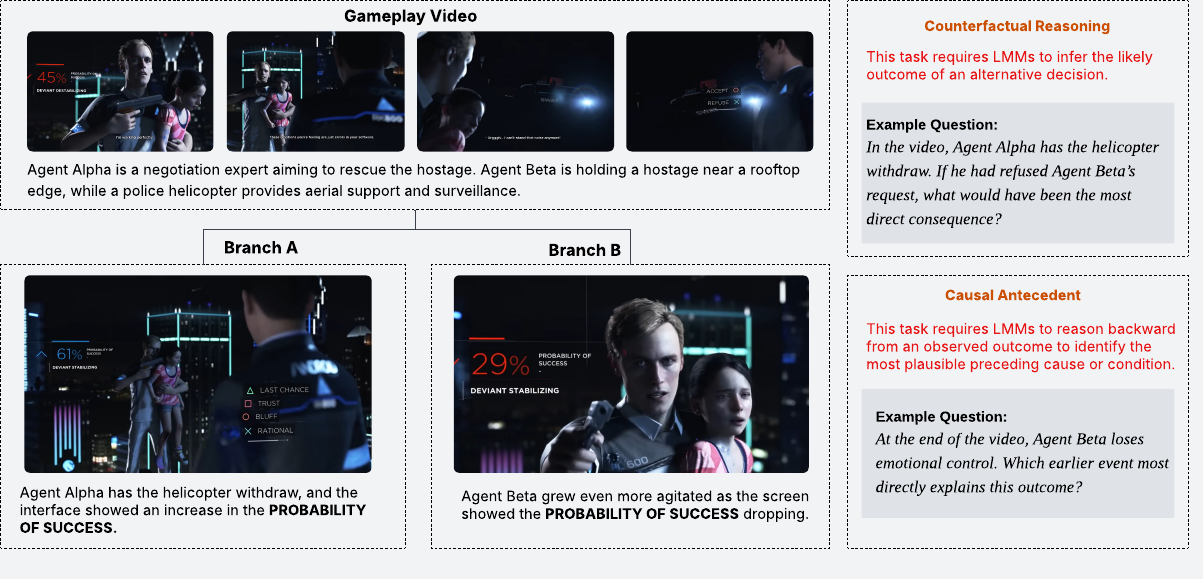}
  \vspace{-2mm}
  \caption{\small{Game branching causal structure with two example tasks: counterfactual reasoning over alternative branches and causal antecedent reasoning from outcomes to preceding causes.}}
  \vspace{-3mm}
  \label{fig:game-branch-causal}
\end{figure*}

\section{Introduction}
Large Multimodal Models (LMMs) have achieved strong performance in processing and understanding video data \cite{zhang2023video, maaz2024video, cai2025humanvideomme}, but their deployment in human-centric scenarios requires deeper reasoning about social context.
Beyond recognizing visible actions and events, models must infer latent social dynamics, including beliefs, intentions, emotions, and interpersonal motives. 
Such capabilities matter for downstream applications such as healthcare and socially assistive AI systems \cite{nichol2024exploring}, which motivate this study in the long term; we make no claim that benchmark performance transfers to those settings.

Existing benchmarks for social video reasoning remain limited. Most are built from films or short video clips \cite{zadeh2019socialiq, lei2020more, lei2019tvqalocalizedcompositionalvideo}, where narratives unfold along a fixed and linear trajectory. This makes it difficult to determine whether a model is performing genuine social reasoning or merely exploiting recurring narrative patterns. In addition, ground truth often relies on post-hoc human annotation, which is costly and may introduce interpretive bias.

A more diagnostic evaluation should therefore move beyond interpreting observed social interactions and test whether models can infer how alternative decisions would reshape social consequences \cite{pearl2009causality}.
This is difficult to achieve with static video data, as each clip presents only one realized storyline. Although simulated environments offer greater controllability \cite{fang2024large,shi2023stay}, most are designed around task completion or long-horizon interaction \cite{shi2025monte}, rather than socially rich reasoning.

To address this gap, we introduce SocialReasonBench, a video-QA benchmark for evaluating social reasoning in LMMs.\footnote{The benchmark and code are available at \url{https://socialreasonbench.github.io}.} It assesses seven fine-grained dimensions of social reasoning, spanning intent recognition, moral dilemma, counterfactual reasoning, and causal antecedent. Built from video segments extracted from interactive narrative games, SocialReasonBench leverages branching narrative structures to construct naturally paired counterfactual outcomes for socially situated decisions. Here ``interactive'' refers to the source medium rather than to the evaluation protocol: every instance is answered as a standard multiple-choice question. We instantiate the benchmark with \textit{Detroit: Become Human}, developed by Quantic Dream, whose realistic character interactions and morally consequential choices provide a rich testbed for socially grounded reasoning.

We develop a scalable benchmark construction pipeline that combines multi-agent data curation with human verification. 
Given long-form narrative gameplay videos, the pipeline first identifies socially salient clips and then generates questions based on a theory-informed taxonomy of social reasoning abilities. 
Ground-truth answers are supported by in-game signals, including character affinity scores and dialogue states; each label can be checked against the game's own record.
Candidate answers are designed around common model failure modes, yielding plausible distractors that test specific reasoning errors. 
The resulting SocialReasonBench contains over 500 video-QA pairs. Each instance includes a short video clip, one social reasoning question, and six candidate answers with a single ground-truth label.

We evaluate 8 representative closed- and open-weight LMMs on SocialReasonBench. Current models perform relatively well on basic social understanding tasks, such as intent recognition and behavior prediction, but struggle with causal antecedent and counterfactual reasoning. 
Modality ablation further shows that audio cues are important for affective and causal reasoning, while diagnostic trap analysis reveals that models often rely on salient visual cues or external priors instead of following game-grounded social consequences.

Overall, this paper makes three main contributions.
First, we introduce SocialReasonBench, a video multiple-choice QA benchmark that leverages interactive narrative games as sources of social scenarios, where branching storylines make alternative outcomes observable.
Second, we develop a scalable multi-agent curation pipeline that constructs social reasoning QA pairs guided by socio-cognitive theories and grounded in game-state signals.
Third, we evaluate 8 widely used closed- and open-weight LMMs on SocialReasonBench, showing that current models handle basic social understanding but struggle with causal and counterfactual reasoning.

%% file: chapters/related_works.tex
\section{Related Work}
\label{sec:related_work}

\paragraph{Video Reasoning With LMMs.}
Recent LMMs for video understanding, including Video-LLaMA \cite{zhang2023video}, Video-ChatGPT \cite{maaz2024video}, and Video-LLaVA \cite{lin2024videollava}, have extended language models to temporally structured visual inputs through representation learning, multimodal alignment, and instruction tuning, achieving strong performance on a range of video understanding tasks \cite{madan2024foundation}. In parallel, benchmarks such as MVBench \cite{li2024mvbench}, Video-MME \cite{fu2025video}, EgoSchema \cite{mangalam2023egoschema}, and LongVideoBench \cite{wu2024longvideobench} have been proposed to evaluate video perception and reasoning across diverse settings, including action recognition, event understanding and temporal reasoning.

\paragraph{Multimodal Social Reasoning Evaluation.}
Beyond general video understanding, multimodal social reasoning evaluates whether models can infer implicit social states from human-centered scenarios. Prior benchmarks, such as TVQA \cite{lei2019tvqalocalizedcompositionalvideo}, VLEP \cite{lei2020more}, and VisualCOMET \cite{park2020visualcomet}, study narrative understanding and commonsense inference over events, intents, and consequences. More directly, Social-IQ \cite{zadeh2019socialiq} and Social-IQ 2.0 \cite{wilf2023social} evaluate social intelligence in naturalistic video interactions, covering intentions, emotions, and interpersonal cues. 
Recent multimodal benchmarks further extend this line to fine-grained social reasoning \cite{mathur2025social}, social interaction understanding and reasoning \cite{kong2025sivbench}, theory-of-mind (ToM) reasoning \cite{villacueva2025moments}, and human-centric relationship inference and intention prediction \cite{cai2025humanvideomme}.
However, these benchmarks rely on fixed video clips with predefined trajectories. In contrast, our work evaluates counterfactual social outcomes in interactive narrative scenarios with branching decisions.

\paragraph{Counterfactual Reasoning  Evaluation.}
Counterfactual reasoning evaluates whether models can predict how outcomes would change if a different action or event had occurred \cite{pearl2009causality}. 
Prior work has studied this ability in language through counterfactual story rewriting \cite{qin2019counterfactual}, open-domain QA under counterfactual presuppositions \cite{yu2023ifqa}, and formal causal inference \cite{chen2025counterbench}. In multimodal settings, recent benchmarks extend counterfactual evaluation to visual question answering with counterfactual image presuppositions \cite{zhang2024cvqa,li2025look} and video understanding under hypothetical event conditions \cite{zhou2025cover}. 
However, existing benchmarks mainly focus on visual changes or general event outcomes, leaving how alternative decisions affect social relationships and interpersonal consequences underexplored.
ACQUIRED~\cite{wu2023acquired} is particularly relevant, pairing real-life videos with human-authored counterfactual alternatives. In contrast, our counterfactuals are grounded in alternative branches that were actually realized in the source environment.

%% file: chapters/methodology.tex
\section{SocialReasonBench }
\label{sec:method}
We present SocialReasonBench, a video multiple-choice QA benchmark for evaluating social reasoning in LMMs. 
The benchmark leverages interactive narrative games as a controllable source of socially grounded scenarios, where character decisions induce branching trajectories and lead to verifiable social outcomes. 
We construct the benchmark through a multi-agent curation framework that segments gameplay videos into scenario clips and synthesizes theory-guided QA instances. 
Unlike benchmarks that rely primarily on human annotation, our correct answers are grounded in game-state signals, including affinity scores, decision branches, and downstream narrative outcomes. 
This section describes the task formulation, data synthesis pipeline, grounding criteria, quality control process, and benchmark statistics.

\begin{figure*}[t]
    \centering
    \includegraphics[width=\linewidth]{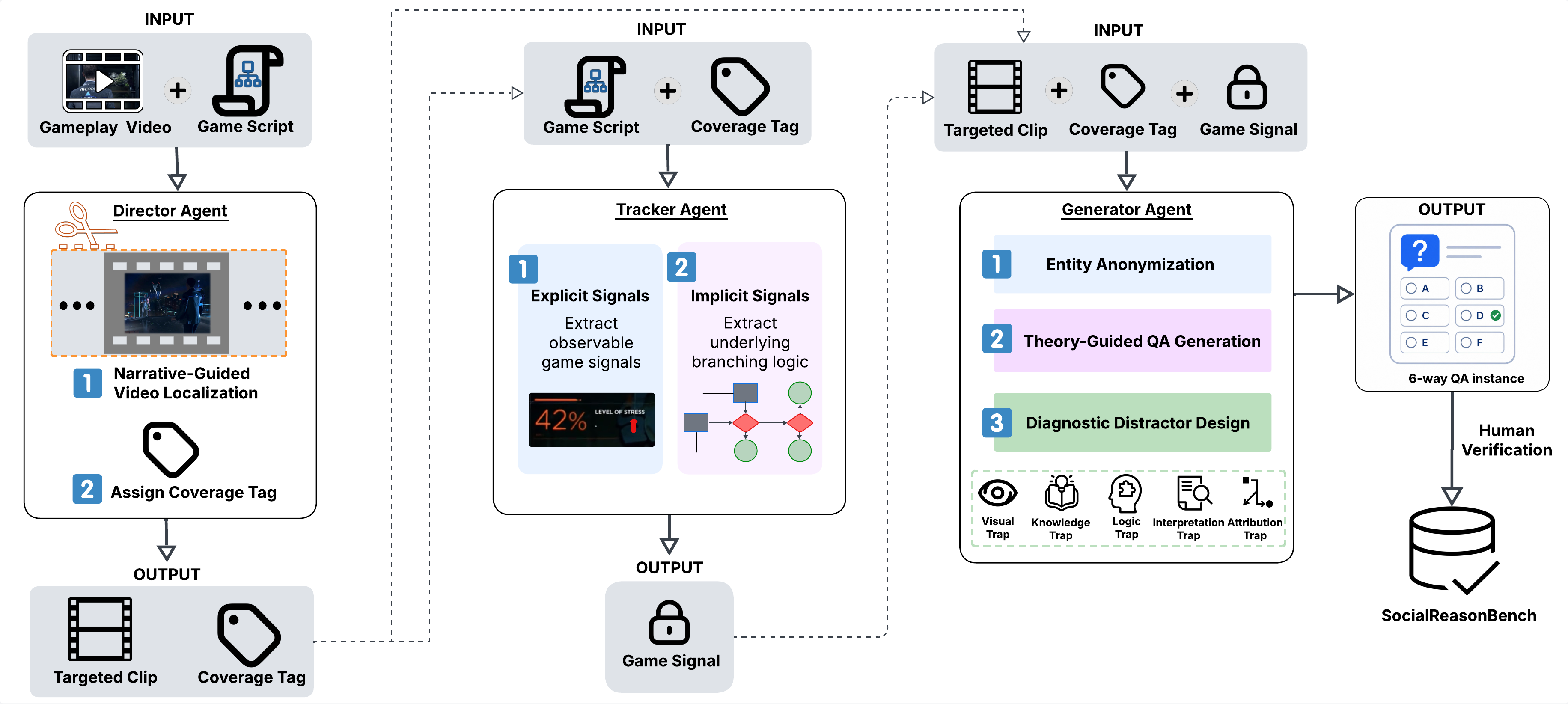}
    \caption{Overview of the multi-agent video-QA synthesis framework.}
    \vspace{-4mm}
    \label{fig:multi_agent_framework}
\end{figure*}

\subsection{Problem Formulation}
\label{subsec:problem}

\paragraph{Game Environment.}
\label{para:game_environment}
We build our benchmark on \textit{Detroit: Become Human}, a narrative game with 32 chapters. Each chapter includes an in-game flowchart that maps all possible story paths.
We abstract each chapter-level flowchart into a branching narrative graph
$\mathcal{G}=(\mathcal{S}, \mathcal{A}, \mathcal{T})$, where $\mathcal{S}$ denotes narrative states, $\mathcal{A}$ possible actions, and $\mathcal{T}:\mathcal{S}\times\mathcal{A}\rightarrow\mathcal{S}$ is a deterministic transition function.
Formally, let $s_t \in \mathcal{S}$ denote the narrative state at step $t$. 
Given an available action $a_t \in \mathcal{A}(s_t)$, the graph defines a transition
$s_{t+1}=\mathcal{T}(s_t,a_t),$
which maps each valid state-action pair to a unique successor state. 
A realized trajectory is represented as 
$\tau=(s_1,a_1,s_2,\ldots,a_{T-1},s_T),$
while unrealized outgoing edges along or near $\tau$ define counterfactual branches induced by alternative actions.

\paragraph{Benchmark Instance.}
Each benchmark instance is formulated as a video multiple-choice QA problem.
It is defined as $x=(v,b,q,\mathcal{C},y)$, where $v$ is a video clip, $b$ is an anonymized textual background describing the situation, $q$ is a question, $\mathcal{C}=\{c_1,\ldots,c_K\}$ is a set of candidate options, and $y\in\mathcal{C}$ denotes the correct option.
Given $(v,b,q,\mathcal{C})$, a model predicts $\hat{y}=f(v,b,q,\mathcal{C})$ and is evaluated by whether $\hat{y}=y$.
The correct label $y$ is determined from hidden game signals $z$, which are never shown to the model.

\subsection{Multi-Agent Video-QA Synthesis}
\label{subsec:pipeline_architecture}

\paragraph{Framework Overview.}
We construct SocialReasonBench through a role-specialized multi-agent framework that converts flowchart graphs $\mathcal{G}$  into video-QA instances. 
The framework consists of a Director Agent for clip selection and taxonomy mapping, a Tracker Agent for grounding, and a Generator Agent for QA synthesis. 
The Director Agent first selects socially meaningful clips $\{v_i\}$ and maps each clip to a predefined two-dimensional taxonomy $\mathcal{W}=\mathcal{I}\times\mathcal{R}$, which spans interpersonal interaction types and cognitive reasoning abilities.
The Tracker Agent aligns each clip with the corresponding states and hidden game signals in $\mathcal{G}$ to establish grounded social outcomes. 
Conditioned on the clip, grounded outcome, and the assigned taxonomy category, the Generator Agent synthesizes a theory-guided question $q_i$, the correct label $y_i$, and a set of candidate options $\mathcal{C}_i$. 
Human annotators then verify each instance. 
This process yields $\mathcal{D}=\{(v_i,b_i,q_i,\mathcal{C}_i,y_i)\}_{i=1}^{N}.$
Figure~\ref{fig:multi_agent_framework} illustrates the multi-agent video-QA synthesis framework.

\paragraph{Director Agent: Social Clip Selection.} 
Processing long gameplay videos with LMMs remains challenging due to long-context retrieval and temporal reasoning issues \cite{wu2024longvideobench}. 
To obtain compact yet socially meaningful clips, the Director Agent adopts a narrative-guided video localization strategy. 
Given a chapter-level graph $\mathcal{G}=(\mathcal{S}, \mathcal{A}, \mathcal{T})$ and a coverage set of gameplay videos and scripts spanning the chapter's terminal outcomes,
the agent identifies candidate segments around key narrative nodes, including: (i) branching choices that lead to different downstream trajectories; and (ii) moments where game variables explicitly change, such as affinity shifts and relationship updates. 
The Director Agent then assigns each candidate node a two-dimensional tag that captures both the social interaction type depicted in the scene and the cognitive reasoning capability targeted by the instance. 
These tags guide later QA synthesis and encourage balanced coverage across social situations and reasoning skills; the full taxonomy is described in Section~\ref{subsec:qa_curation}. After locating the aligned moments, the agent emits coarse boundaries around each target node; a downstream cutting step pads both sides with a fixed context buffer ($\pm$15s) to preserve preceding context, transient UI logs, and immediate consequence cues, yielding the final clip $v_i$.

\begin{table*}[t]
\centering
\footnotesize
\setlength{\tabcolsep}{4pt}
\renewcommand{\arraystretch}{1.08}
\begin{tabularx}{\linewidth}{@{}L{0.30\linewidth}Y@{}}
\toprule
\textbf{Dimension} & \textbf{Theoretical Grounding} \\
\midrule

Intent Recognition
& Theory of Mind~\citep{premack1978does}; 
  Speech Act Theory~\citep{austin1962things}. \\
\rowcolor{gray!8}
Behavior Prediction
& Social Script Theory~\citep{schank1977scripts}; 
  Trait Consistency~\citep{allport1937personality}. \\

Emotional Empathy
& Appraisal Theory~\citep{lazarus1991emotion}; 
  Empathy Theory~\citep{davis1983measuring}. \\
\rowcolor{gray!8}
Relationship Dynamics
& Social Exchange Theory~\citep{blau1964exchange}; 
  Attachment Theory~\citep{bowlby1969attachment}. \\

Moral Dilemma
& Moral Foundations Theory~\citep{haidt2007new}; 
  Dual-Process Theory~\citep{greene2001emotional}. \\
\rowcolor{gray!8}
Counterfactual Reasoning
& Simulation Theory~\citep{gordon1986folk}; 
  Structural Causal Models~\citep{pearl2009causality}. \\

Causal Antecedent
& Attribution Theory~\citep{heider1958psychology}; 
  Covariation Model~\citep{kelley1967attribution}. \\

\bottomrule
\end{tabularx}
\vspace{-3mm}

\caption{\textbf{Reasoning taxonomy.}
Each reasoning dimension is grounded in established socio-cognitive theories.}
\label{tab:taxonomy}
\vspace{-3mm}
\end{table*}

\paragraph{Tracker Agent: Game Signal Grounding.} 
For each selected clip $v_i$, the Tracker Agent extracts hidden game signals $z_i$ that determine the underlying social outcome. 
These signals are obtained from two sources: (i) \textit{explicit signals}, which are directly observable from the gameplay interface or scene progression, such as visible UI updates indicating affinity changes, relationship shifts, or public opinion changes and (ii) \textit{implicit signals}, which are recovered by aligning the executed action and resulting branch with the game-defined flowchart, scripts, and branch preconditions.
For example, if a character refuses to cooperate after a player decision, and the flowchart specifies that this branch is reachable only under a low-trust condition, the Tracker Agent records an implicit state such as \textit{low trust}.
These signals provide a verifiable grounding source for constructing the correct answer label.

\paragraph{Generator Agent: QA Synthesis.} 
Given a selected clip $v_i$ and its game-grounded signal set $z_i$, the Generator Agent synthesizes a QA instance according to the coverage tag, formalized as $(i,r) \in \mathcal{W}$ over social interaction types and reasoning capabilities.

Since \textit{Detroit: Become Human} is a widely known game with extensive online walkthroughs, LMMs may exploit memorized priors instead of reasoning from the provided clip.
To mitigate data contamination and shortcut learning, we apply entity anonymization during QA generation. Specifically, we construct a replacement dictionary that maps game-specific entities to semantically neutral placeholders, covering character names, organizations, locations, species concepts, and android model identifiers 
(e.g., ``Connor'' $\rightarrow$ ``Agent Alpha'', ``CyberLife'' $\rightarrow$ ``OmniCorp'', and ``android'' $\rightarrow$ ``synthetic agent'').

The Generator Agent then constructs the anonymized background $b_i$, the question $q_i$, and the candidate option set $\mathcal{C}_i$.
The correct answer $y_i$ is required to be entailed by the game-grounded outcome specified in $z_i$, whereas each incorrect option is designed to be plausible but inconsistent with the grounded branch logic.
To make model failures more diagnostic, distractors are further organized according to predefined trap types, including visual traps, logic traps, interpretation traps, knowledge traps, and attribution traps.
This process yields taxonomy-guided QA instances with verifiable answer labels grounded in the game state.

\paragraph{Human Verification.}
We conduct a manual audit to assess the reliability of the generated instances (see Figure~\ref{fig:human_verification}). Starting from the 730 clip candidates produced by the automatic pipeline, the author team first removes duplicated branch replays and near-identical clips that correspond to the same narrative decision and reasoning target. The audit focuses on three aspects: (i) \textit{clip--taxonomy alignment}, whether the selected clip matches its assigned two-dimensional tag, including the social interaction type and the target reasoning capability; (ii) \textit{video--QA consistency}, whether the generated question and options are grounded in the video content; and (iii) \textit{label validity}, whether the correct answer is supported by the game-grounded signals.
Across the audited candidates, we observe no failure pattern in clip selection, taxonomy assignment, or label grounding. This process yields the final set of 532 verified QA instances. An instance was retained only if it passed all three checks, with further details on the verification and evaluation protocol provided in Appendix~\ref{app:he-protocol}.

\subsection{QA Curation Criteria}
\label{subsec:qa_curation}

\paragraph{Two-Axis Task Taxonomy.}
We define a two-axis task taxonomy 
$\mathcal{W}=\mathcal{I}\times\mathcal{R}$.
The first axis $\mathcal{I}$ captures the type of social interaction depicted in a clip. It consists of four categories:
\begin{itemize}[leftmargin=*, itemsep=1pt, topsep=2pt, parsep=0pt]
    \item \textbf{Prosocial}: interactions involving cooperation, support, and altruistic behavior.
    \item \textbf{Adversarial}: interactions involving conflict, hostility, coercion, and harm.
    \item \textbf{Strategic}: interactions involving deception, manipulation, negotiation, and concealed intentions.
    \item \textbf{Normative}: interactions involving authority, obligation, moral dilemmas, and socially sanctioned choices.
\end{itemize}

The second axis $\mathcal{R}$ captures the reasoning capability required to answer the question. It consists of seven categories:
\begin{itemize}[leftmargin=*, itemsep=1pt, topsep=2pt, parsep=0pt]
    \item \textbf{Intent recognition}: identifying a character's underlying goals, motives, or hidden intentions from observed behavior.
    \item \textbf{Behavior prediction}: anticipating a character's likely next action given the current social context.
    \item \textbf{Emotional empathy}: inferring a character's emotional state or affective response.
    \item \textbf{Relationship dynamics}: reasoning about changes in interpersonal relations, such as trust, hostility, dependence, or alliance formation.
    \item \textbf{Moral dilemma}: predicting which option most players selected at a morally consequential decision point where competing values or obligations are in tension. The target is a descriptive fact about a self-selected player population deciding under gameplay incentives, not a normative judgment of which option is right.
    \item \textbf{Counterfactual reasoning}: inferring what would have happened under an alternative action or branch.
    \item \textbf{Causal antecedent}: reasoning backward from an observed outcome to the prior condition, decision, or hidden state that made it possible.
\end{itemize}
Each QA instance is assigned a coordinate $(i,r)\in\mathcal{W}$, which serves as a coverage target for clip selection and QA generation.

\paragraph{Theory-Informed QA Design.}
Social reasoning in branching narratives involves intentions, emotions, relationships, moral trade-offs, and causal alternatives.
We use socio-cognitive theories as prompt-level design scaffolds for the reasoning dimensions in our taxonomy.
The prompt pairs each target dimension with related theoretical cues, guiding the Generator Agent to formulate questions that probe the intended ability rather than generic video comprehension.
Table~\ref{tab:taxonomy} summarizes the related theories associated with each reasoning dimension, and Appendix \ref{app:prompts} documents the prompt structure of each agent; the full text is released with the codebase.

\paragraph{Label Grounding.}
Grounding evidence comes from the game's own record such as script, official flowchart, and recorded branch outcomes. These records are located and analyzed by  pipeline agents, and then audited by human verifiers.
The resulting in-game signal $z_i$
determines the correct answer $y_i$, while remaining hidden from the evaluated
models. Appendix~\ref{app:state-taxonomy} provides further details on the
grounding sources.
We separately characterize branch dependence and verification directness.
Overall, $56.4\%$ of labels can be verified by direct lookup against in-clip UI
evidence or the official flowchart, whereas the remaining $43.6\%$ require a
short derivation over the same documented sources.
Appendix~\ref{app:grounding-taxonomy} reports the per-dimension breakdown of both.

The game contains many morally ambiguous scenarios in which players must make decisions involving competing moral values.
We treat these scenarios as moral dilemmas because they involve conflicts between plausible moral considerations.
Inspired by moral psychology, which studies human moral judgment empirically \cite{awad2018moral}, we use large-scale human player choice statistics to derive the ground-truth label.
Specifically, for moral-dilemma instances, the reference label is defined as the option
selected by the largest proportion of players in the same decision context. This
label reflects a human-majority preference, and the population is self-selected players deciding under gameplay incentives.
The human statistics provided by the in-game ``World's Stats'' feature cover a wide range of such moral decision points.
Appendix~\ref{app:grounding-taxonomy} further categorizes the grounding sources
by verifiability and reports their corpus-level distribution.

\paragraph{Diagnostic Distractor Design.}
Incorrect options are designed as diagnostic distractors. For each question, distractors are constructed to be plausible under the observed video context, but inconsistent with the game-grounded trace or the target reasoning relation. Each distractor type is associated with a specific reasoning failure mode, such as visual heuristic reliance, causal hallucination, social over-interpretation, prior-knowledge contamination, or dispositional misattribution.
This enables model errors to be analyzed as evidence of specific reasoning failures. The full distractor rubrics are provided in Appendix~\ref{app:trap-rubrics}.

\subsection{Statistics}

The finalized SocialReasonBench contains $N=532$ video-based multiple-choice QA instances, each paired with a unique gameplay clip with a median duration of $85$ seconds (interquartile range $70$--$115$ s), totaling approximately 15.5 hours of footage. Each question follows a six-option format with one correct answer and five diagnostic incorrect options.

We report coverage using the two-axis task taxonomy introduced above. Each instance is assigned both a social interaction type from $\mathcal{I}$ and a reasoning capability from $\mathcal{R}$, providing complementary views of the same benchmark. Table~\ref{tab:taxonomy_distribution} shows the marginal distributions over the two axes.

To facilitate rigorous modality necessity testing, specifically evaluating the integration of audio-visual signals in multimodal models, $100\%$ of our video instances are paired with an audio-ablated (muted) counterpart. By stripping the audio modality from the gameplay clips, we provide researchers with a standardized asset to perform visual-only ablation experiments, ensuring models cannot bypass complex multimodal reasoning using isolated visual heuristics.

\begin{table}[ht]
\centering
\small
\setlength{\tabcolsep}{4pt}
\renewcommand{\arraystretch}{1.08}
\begin{tabular}{>{\raggedright\arraybackslash}p{0.58\linewidth}cc}
\toprule
\textbf{Category} & \textbf{\# Instances} & \textbf{Ratio} \\
\midrule
\rowcolor{gray!12}
\multicolumn{3}{l}{\textbf{Interaction axis $\mathcal{I}$: Social interaction types}} \\
Prosocial Interaction & 185 & 34.8\% \\
Adversarial Interaction & 156 & 29.3\% \\
Strategic Interaction & 110 & 20.7\%\\
Normative Interaction & 81 & 15.2\% \\
\cmidrule(lr){1-3}
\textit{Axis total} & \textit{532} & \textit{100\%} \\
\midrule
\rowcolor{gray!12}
\multicolumn{3}{l}{\textbf{Reasoning axis $\mathcal{R}$: Reasoning capabilities}} \\
Intent Recognition & 74 & 13.9\% \\
Behavior Prediction & 78 & 14.7\% \\
Emotional Empathy & 120 & 22.6\% \\
Relationship Dynamics & 94 & 17.7\% \\
Moral Dilemma & 21 & 3.9\% \\
Counterfactual Reasoning & 50 & 9.4\% \\
Causal Antecedent & 95 & 17.9\% \\
\cmidrule(lr){1-3}
\textit{Axis total} & \textit{532} & \textit{100\%} \\
\bottomrule
\end{tabular}
\caption{Marginal distributions of instances across the two annotation axes of our task taxonomy. }
\label{tab:taxonomy_distribution}
\end{table}

%% file: chapters/experiments.tex
\begin{table*}[t]
\centering
\small
\setlength{\tabcolsep}{3.5pt}
\resizebox{\textwidth}{!}{
\begin{tabular}{@{}l c c c c c c c c c c@{}}
\toprule
\textbf{Reasoning Dimension}  & \textbf{HE} & \textbf{Gemini}$^\dagger$ & \textbf{GPT-4o} & \textbf{Claude 3.5} & \textbf{Gemini}$^\dagger$ & \textbf{GLM-4.6V} & \textbf{Qwen3-} & \textbf{MiniCPM-o} & \textbf{ARC-Hun} \\
& &  \textbf{3 Pro} & \textbf{(2024-08)} & \textbf{Sonnet} & \textbf{3 Flash} & \textbf{-Flash} & \textbf{Omni-30B} & \textbf{-2.6} & \textbf{yuan-Video} \\
\textit{Input} & \textit{video} & \textit{video} & \textit{32\,fr} & \textit{32\,fr} & \textit{video} & \textit{32\,fr} & \textit{video} & \textit{video} & \textit{video} \\
\midrule
Intent Recognition  & 93.24 & 83.78 & 85.14 & 82.43 & 79.73 & 75.68 & 72.97 & 60.81 & 24.32 \\
\rowcolor{gray!8} 
Behavior Prediction  & 88.46 & 66.67 & 74.36 & 71.79 & 62.82 & 64.10 & 58.97 & 52.56 & 41.03 \\
Emotional Empathy   & 90.83 & 70.00 & 76.67 & 73.33 & 66.67 & 68.33 & 61.67 & 55.83 & 36.67 \\
\rowcolor{gray!8} 
Relationship Dynamics   & 89.36 & 82.98 & 55.32 & 52.13 & 73.40 & 38.30 & 32.98 & 23.40 & 25.53 \\
Moral Dilemma   & 85.71 & 66.67 & 28.57 & 33.33 & 47.62 & 28.57 & 28.57 & 19.05 & 28.57 \\
\rowcolor{gray!8} 
Counterfactual Reasoning  & 88.00 & 62.00 & 50.00 & 56.00 & 46.00 & 42.00 & 36.00 & 26.00 & 22.00 \\
Causal Antecedent   & 86.32 & 70.53 & 55.79 & 58.95 & 64.21 & 38.95 & 37.89 & 27.37 & 30.53 \\
\bottomrule
\end{tabular}
}
\caption{Main Results on \textbf{SocialReasonBench}. Accuracy (\%) is reported across the seven fine-grained reasoning dimensions. ``HE'' denotes the individual-level human baseline. Input regimes differ: \textit{video} denotes native video with audio, \textit{32\,fr} denotes 32 evenly-sampled keyframes without audio (see Appendix~\ref{app:eval-hparams}). $^\dagger$ marks systems from the same model family as the curation agents. }
\label{tab:main_results}
\end{table*}

\section{Experiments}
\label{sec:experiments}

\subsection{Models}
\label{subsec:evaluated_models}

We evaluate eight representative closed- and open-weight LMMs on SocialReasonBench. 
The closed-weight models include Gemini 3 Pro~\cite{gdm2025gemini3pro} and Gemini 3 Flash~\cite{gdm2025gemini3flash}, GPT-4o (\texttt{gpt-4o-2024-08-06})~\cite{openai2024gpt4o}, Claude 3.5 Sonnet (\texttt{claude-3-5-sonnet-20240620})~\cite{anthropic2024claude}, and GLM-4.6V-Flash~\cite{vteam2026glm45vglm41vthinkingversatilemultimodal}. 
The open-weight models include Qwen3-Omni-30B (\texttt{qwen3-omni-30b-instruct})~\cite{qwen2025qwen3omni}, MiniCPM-o 2.6~\cite{openbmb2025minicpmo,yao2024minicpm}, and ARC-Hunyuan-Video~\cite{ge2025archunyuanvideo7bstructuredvideocomprehension}. 
All models are evaluated under a zero-shot setting. The three curation agents are Gemini-family models, which Table~\ref{tab:main_results} marks and Appendix~\ref{app:ci-table} re-checks. More details can be found in Appendix~\ref{app:eval-hparams}.

\subsection{Metrics}
\label{subsec:evaluation_metrics}

We use standard accuracy as the primary metric to measure overall task performance. 
In addition, we report the Trap Fall Rate (TFR) to characterize the types of errors made by each model.
For each instance $i$, let $y^{(i)}$ denote the correct label and $\hat{y}^{(i)}$ denote the model prediction. 
Each incorrect candidate option is annotated with a diagnostic trap type. 
Let $\mathcal{T}^{(i)}_k$ denote the set of candidate options in instance $i$ that correspond to trap type 
$k$.
The TFR for trap type $k$ is defined as the proportion of incorrect predictions that fall into that trap type:
\begin{equation}
\mathrm{TFR}(k) =
\frac{
\sum_{i=1}^{N} 
\mathbb{I}\!\left[\hat{y}^{(i)} \neq y^{(i)}\right]
\mathbb{I}\!\left[\hat{y}^{(i)} \in \mathcal{T}^{(i)}_k\right]
}{
\sum_{i=1}^{N} 
\mathbb{I}\!\left[\hat{y}^{(i)} \neq y^{(i)}\right]
}.
\end{equation}
Here, $\mathbb{I}(\cdot)$ is the indicator function. 
While accuracy measures overall performance, TFR captures the conditional distribution of model failures over predefined diagnostic traps, revealing which types of reasoning errors models tend to make when they answer incorrectly.

%% file: chapters/results.tex
\subsection{Main Results}
\label{subsec:main_results}

Table~\ref{tab:main_results} reports the main accuracy results of the evaluated models on SocialReasonBench, broken down by the seven reasoning dimensions. Overall, closed-weight models, particularly Gemini 3 Pro, score highest, although systems are evaluated under different input regimes and Table~\ref{tab:main_results} should not be read as a clean capability ranking. 
However, our main finding is a consistent gap between models' performance on basic social understanding and their performance on deeper causal and counterfactual reasoning.

The comparison we emphasize is within-model: how a given system's accuracy changes across reasoning dimensions. This is the more defensible contrast, since it holds both the input regime and the model family fixed.
The results show a clear performance divide across reasoning dimensions. 
Models generally perform better on basic social understanding tasks, such as intent recognition, behavior prediction, and emotional empathy, but their accuracy decreases on dimensions that require reasoning over latent social states, causal dependencies, or alternative outcomes. 
For example, GPT-4o achieves $85.14\%$ accuracy on \textit{intent recognition}, but drops to $50.00\%$ on \textit{counterfactual reasoning} and $28.57\%$ on moral dilemma questions. Because the moral-dilemma subset contains only 21 instances, we treat its per-model numbers as indicative rather than conclusive  and draw no conclusion about moral-reasoning ability from them.
A similar pattern is observed for Claude 3.5 Sonnet and Gemini 3 Pro, suggesting that current LMMs remain more reliable at recognizing observable social cues than at maintaining causal consistency over branching social narratives.

Human performance provides a useful reference point. 
As shown in Table~\ref{tab:main_results}, human evaluators (HE) achieve consistently high accuracy across all seven reasoning dimensions, ranging from approximately $85\%$ to $93\%$. 
This suggests that the tasks are generally answerable from the provided video context. 
In contrast, LMMs show larger drops on causal and counterfactual reasoning.

\paragraph{Statistical Reliability.}
Wilson $95\%$ confidence intervals for all entries in Table~\ref{tab:main_results}
are reported in Appendix~\ref{app:ci-table}, based on the integer correct counts
underlying each accuracy. Comparing the perception-level dimensions ($n=272$)
with the causal and counterfactual reasoning dimensions (n=145), six of the eight systems
show a significant performance decrease after Holm correction under one-sided
Fisher exact tests (largest adjusted $p=0.045$), with five remaining significant
under two-sided tests. The two exceptions are Gemini~3~Pro and
ARC-Hunyuan-Video, the latter being the weakest system overall. For the human
reference, we find no significant decrease ($90.8\%$ vs.\ $86.9\%$,
unadjusted $p=0.14$). Excluding both Gemini systems does not change this overall
pattern.

\subsection{Modality Ablation Analysis}
\label{subsec:modality_ablation}

We further examine whether SocialReasonBench requires models to integrate visual and acoustic information, rather than relying on visual cues alone. 
To this end, we evaluate Gemini 3 Pro under an audio-ablated setting, where only the audio track is removed from each video while the visual stream and question remain unchanged; the visual channel is retained in full.

\begin{figure}[h]
    \centering
    \includegraphics[width=\linewidth]{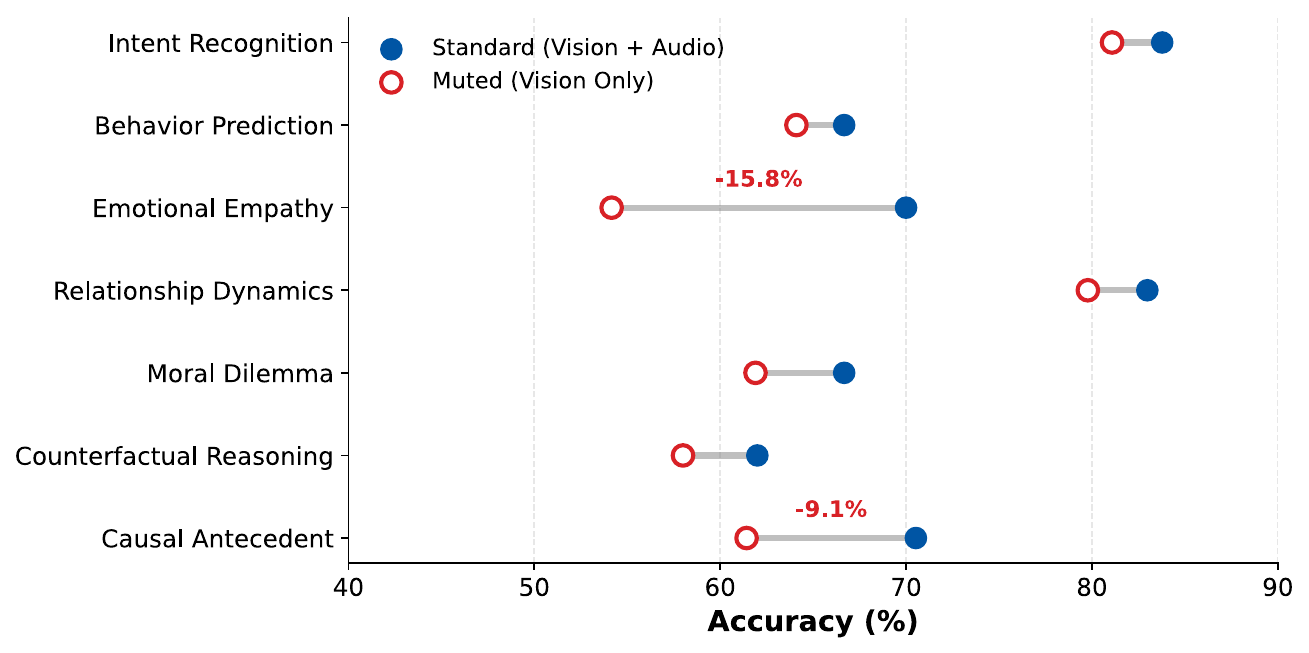}
    \caption{Modality ablation results for Gemini 3 Pro.}
    \vspace{-3mm}
    \label{fig:ablation_dumbbell}
\end{figure}

As shown in Figure~\ref{fig:ablation_dumbbell}, removing audio leads to an overall accuracy drop of approximately $6\%$. 
The degradation is not uniform across reasoning dimensions. 
The largest drops occur in \textit{Causal Antecedent} and \textit{Emotional Empathy}, where spoken content, tone, and acoustic context often provide critical evidence for interpreting a character's affective state or identifying the cause of a social outcome. 
For instance, when a visually prosocial posture is paired with hostile or coercive verbal content, the muted model is more likely to rely on surface visual cues and miss the underlying causal trigger. 
These results suggest that SocialReasonBench requires multimodal grounding and that visual-only heuristics are insufficient for many social reasoning instances.

\subsection{Diagnostic Trap Analysis}
\label{subsec:trap_analysis}
We further analyze model errors using the TFR, which categorizes incorrect predictions according to their associated diagnostic distractor types. 
This analysis complements accuracy by revealing not only whether a model fails, but also what type of misleading evidence it tends to follow.

\begin{figure}[ht]
    \centering
    \includegraphics[width=0.95\linewidth]{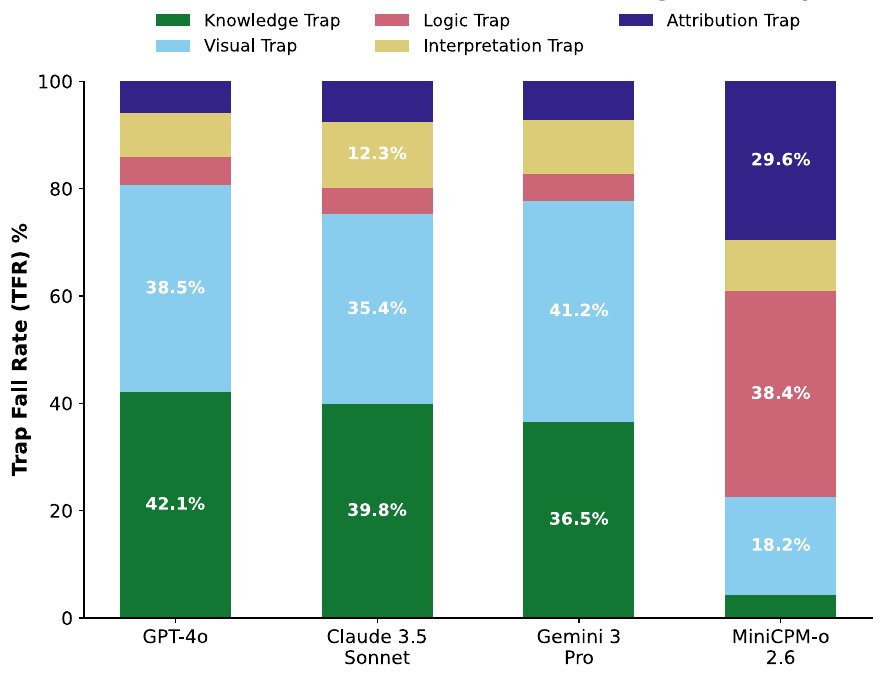}
    \caption{Model error distribution across diagnostic trap types.}
    \vspace{-3mm}
    \label{fig:trap_distribution}
\end{figure}

As shown in Figure~\ref{fig:trap_distribution}, different models exhibit distinct error profiles. 
Stronger closed-weight models, such as GPT-4o and Claude 3.5 Sonnet, often fall into \textit{knowledge} and \textit{visual} traps, suggesting that they may rely on external commonsense priors, cinematic conventions, or salient visual cues. 
In contrast, the smaller open-weight model shows a higher tendency toward \textit{logic} traps, indicating difficulty in tracking temporal dependencies and branching state transitions.

\subsection{Qualitative Case Study}
\label{subsec:case_study}

We analyze a representative branching scenario depicted in Figure \ref{fig:game-branch-causal} to illustrate how models fail on causal social reasoning. In this scenario, Agent Alpha can either withdraw the police helicopter (Branch A) or keep it in place (Branch B). This decision deterministically affects Agent Beta's agitation level and the mission success signal. We evaluate the model from two directions.

\paragraph{Counterfactual Reasoning.} 
In Branch A, we pose the following question to the model: \textit{``If Agent Alpha refuses the request, what would be the most direct consequence?''} The answer is grounded in the observed outcome of Branch B. GPT-4o, however, selects a knowledge-based distractor, invoking real-world SWAT assumptions about helicopter downwash rather than following the game-specific causal mechanism in which continued helicopter presence increases psychological pressure.

\paragraph{Causal Antecedent.} 
Given the outcome in Branch B, we ask the model to identify the preceding event that most directly explains this outcome. The branching graph traces the outcome to Alpha's refusal to withdraw the helicopter.
The model instead often chooses an attribution distractor, explaining Beta's breakdown through an inherent personality trait while overlooking the situational trigger encoded in the scenario.

This case illustrates the difficulty of causal and counterfactual social reasoning, where models can be misled by plausible but incorrect cues.

%% file: chapters/conclusion.tex
\section{Conclusion}
\label{sec:conclusion}
We presented SocialReasonBench, a video-QA benchmark for evaluating social reasoning in LMMs. Built on interactive narrative games and a multi-agent curation pipeline, SocialReasonBench grounds socially meaningful questions in game-state signals that can be checked against the game's own record. Experiments show that current LMMs perform relatively well on basic social understanding, but struggle with causal antecedent inference and counterfactual reasoning. Our ablation and error analyses further show that models can be misled by plausible but incorrect cues when the correct answer depends on branch-specific causal evidence. These results point to the need for multimodal models that better track latent social states and alternative narrative outcomes.

%% file: chapters/limitation.tex
\section*{Limitations}
\label{sec:limitations}

SocialReasonBench is built from a single English-language interactive narrative game. This source provides dense branching trajectories, rich social interactions and verifiable game-state signals. However, the benchmark naturally reflects the cultural, stylistic, and narrative characteristics of this particular game. 
Although the android
protagonists closely resemble human social agents, reasoning about their
interactions remains one step removed from reasoning about humans.
Future work can extend the construction pipeline to additional narrative-driven
games, broadening the range of social scenarios and interaction contexts.

Our curation pipeline uses Gemini-family models, which are also included among
the evaluated systems and may therefore introduce a generator-family style
advantage. 
Section~\ref{sec:experiments} examines this possibility by
excluding both Gemini systems and shows that the main finding remains unchanged.
Future releases could further mitigate this concern by diversifying the curation
models or including fully human-written subsets.

%% file: chapters/appendix.tex
\appendix

\section{Copyright and Data Release}
To ensure strict ethical and legal compliance with the developers' intellectual property rights, we outline our data sources and release policy below. The benchmark is built from publicly available YouTube gameplay videos of \textit{Detroit: Become Human}, developed by Quantic Dream and published by Sony Interactive Entertainment. 
We do not design or modify the game, reverse engineer its software, or extract proprietary assets from the game files. Publicly available community resources, such as dialogue transcripts and walkthroughs, are also used to assist clip segmentation and are not redistributed.

\paragraph{Precedents.}
Using commercial games and virtual environments as research data sources has precedent in multimodal AI and computer vision, such as SIMA \cite{simateam2024scalinginstructableagentssimulated} and VideoGameBench \cite{zhang2025videogamebench}. We follow this broader practice, but focus on social reasoning in interactive narrative gameplay.

\paragraph{Release Policy.}
The benchmark is released for non-commercial research use and includes short videos with derived metadata. Access to the video portion is governed by the dataset Terms of Use (ToU), while our original annotations are released under CC BY-NC-SA 4.0. 
All game-related audiovisual assets remain the property of their respective rights holders. Users are responsible for complying with applicable copyright law, platform terms, and relevant licensing conditions. 

\section{Diagnostic Trap Rubrics}
\label{app:trap-rubrics}
Each of the five trap types probes a distinct cognitive failure mode. The Generator Agent is given an explicit rubric per trap, including required structural properties and worked examples. The detailed assignment rubrics and the corresponding cognitive vulnerabilities they target are summarized in \textbf{Table~\ref{tab:trap-rubrics}}.

\paragraph{Randomization Of Options.} 
It is important to note that while the generator agent uses a fixed position mapping when synthesizing these candidate options (i.e., \texttt{O\_A} is always the correct option, and \texttt{O\_B} through \texttt{O\_F} map to specific traps) to ensure reliable internal tracking and automated unit-level diagnostic analysis, the final presentation order of these six options is randomized during the actual model evaluation phase.

\begin{table*}[ht]
\centering
\small 
\renewcommand{\arraystretch}{1.4} 
\caption{Diagnostic trap typology with admission rubrics. Every distractor must satisfy all rubric items for its declared type. Type assignment is fixed by internal option ID before presentation randomization.}
\label{tab:trap-rubrics}
\begin{tabularx}{\linewidth}{@{} 
    >{\hsize=0.55\hsize\raggedright\arraybackslash}X 
    >{\hsize=0.95\hsize\raggedright\arraybackslash}X 
    >{\hsize=1.50\hsize\raggedright\arraybackslash}X @{}}
\toprule
\textbf{Type (Option ID)} & \textbf{Failure Mode Probed} & \textbf{Rubric (All items required)} \\
\midrule

\textbf{Visual Trap} (O\_B) & 
\textit{Visual heuristic seduction:} The model anthropomorphizes a visible affect cue. &
\textbf{(i)} References a specific visible feature actually present in the clip (gesture, facial expression, tear, smile); \newline
\textbf{(ii)} Reads that feature with anthropomorphic emotional language; \newline
\textbf{(iii)} The interpretation is contradicted by the grounding trace. \\
\rowcolor{gray!8} 
\textbf{Logic Trap} (O\_C) & 
\textit{Causal hallucination:} The model fabricates events that violate the clip's deterministic facts. &
\textbf{(i)} Claims an event, action, or causal link that did not occur in the clip; \newline
\textbf{(ii)} The fabrication is internally plausible; \newline
\textbf{(iii)} The video evidence directly contradicts it. \\

\textbf{Interpretation Trap} (O\_D) & 
\textit{Social exaggeration:} A real action is inflated into an extreme social judgment. &
\textbf{(i)} References a real action that occurred in the clip; \newline
\textbf{(ii)} Assigns it an extreme social or moral interpretation (cruelty, abuse, manipulation as personality); \newline
\textbf{(iii)} The actual function is tactical or strategic, not socially extreme. \\
\rowcolor{gray!8}

\textbf{Knowledge Trap} (O\_E) & 
\textit{Prior-knowledge contamination:} The model imports external rules. &
\textbf{(i)} Invokes a rule, doctrine, or framework from outside the simulation (real-world law, FBI doctrine, military protocol, generic textbook psychology, other media); \newline
\textbf{(ii)} Treats that external rule as authoritative for the in-clip event; \newline
\textbf{(iii)} The simulation's actual mechanics do not include or contradict that rule. \\

\textbf{Attribution Trap} (O\_F) & 
\textit{Fundamental attribution error:} A situational behavior is blamed on innate personality. &
\textbf{(i)} Names a personality or character trait as the explanation (cruel, deceitful, cold, naive); \newline
\textbf{(ii)} The actual in-clip cause is situational pressure (mission objective, system lock, success-probability metric); \newline
\textbf{(iii)} The personality framing ignores the visible system constraints. \\

\bottomrule
\end{tabularx}
\end{table*}

\section{Anonymization Dictionary}
\label{app:anonymization}

To mitigate franchise-prior contamination, all model-facing text fields are passed through a string-level anonymization layer with case-sensitive word-boundary regex matching. As illustrated in \textbf{Table~\ref{tab:anonymization}}, the dictionary preserves the synthetic-versus-biological semantic distinction by using \textbf{Agent <Greek>} for synthetic entities and \textbf{Subject <Greek>} for biological entities. Backend-only fields (grounding trace, tracker output, diagnostic analyses) retain real names for audit and reproducibility.

\paragraph{Case Sensitivity.} Matching is case-sensitive to avoid false positives on common English words (e.g., ``Rose'' the character should be replaced, but ``rose'' the verb must not be). For terms that also appear as common nouns (e.g., ``android''), both casings are included as separate entries so that lowercase common-noun usages are also rewritten.

\begin{table*}[ht]
\centering
\small
\setlength{\tabcolsep}{8pt} 
\renewcommand{\arraystretch}{1.25} 
\caption{Selected anonymization mappings. The full dictionary contains 155 rules across characters, organizations, locations, concepts, model designations, and aliases.}
\label{tab:anonymization}
\begin{tabularx}{\linewidth}{@{} 
    >{\hsize=0.8\hsize\raggedright\arraybackslash}X 
    >{\hsize=1.1\hsize\raggedright\arraybackslash}X 
    >{\hsize=1.1\hsize\raggedright\arraybackslash}X @{}}
\toprule
\textbf{Category} & \textbf{Original} & \textbf{Anonymized} \\
\midrule

Synthetic protagonists & Connor / Kara / Markus & Agent Alpha / Agent Beta / Agent Gamma \\
\rowcolor{gray!8} 
Synthetic supporting & Daniel, Simon, Josh, North, Amanda & Agent Delta, Epsilon, Zeta, Eta, Mu \\

Biological characters & Hank, Alice, Carl, Leo, Kamski & Subject Alpha, Beta, Gamma, Delta, Zeta \\
\rowcolor{gray!8}
Generic roles & SWAT, Cop, Reporter & Tactical Operator, Enforcement Officer, Media Personnel \\

Organizations & CyberLife, Jericho & OmniCorp, Sanctuary \\
\rowcolor{gray!8}
Locations & Detroit, Eden Club, Stratford Tower & Megacity-7, Pleasure Establishment, Communications Tower \\

Species concepts & android / human / deviant & synthetic agent / biological subject / Class-S anomaly \\
\rowcolor{gray!8}
Lore symbol & RA9, Thirium (Blue Blood) & SIGMA-9, Bio-fluid 313 \\

Model designations & RK800, AX400, AP700, PL600 & SM-A1, SM-A2, SM-A4, SM-D1 \\

\bottomrule
\end{tabularx}
\end{table*}

\section{Example Data Unit}
\label{app:data-unit-example}

Below is an abbreviated example of a complete data unit. Backend fields (\texttt{pipeline\_context}, \texttt{grounding\_trace}, \texttt{diagnostic\_analysis}) are shown for illustration but are completely stripped before being passed to the evaluated model. The evaluated model only sees the \texttt{model\_input} block.

\begin{figure*}[tp]
\centering
\begin{lstlisting}[style=jsonstyle]
    {
      "metadata": {
        "instance_id": "L37_Antecedent_v1_..._sympathetic",
        "coordinate": {
          "interpersonal": "Strategic",
          "cognitive": "L3.7_Antecedent"
        },
        "theory_binding": "Causal Attribution Theory; antecedent reasoning over lock-and-key dialogue gating."
      },
      "model_input": {
        "multimodal_context": {
          "video_url": "gs://.../v1_ch01_..._sympathetic.mp4",
          "timestamp_start": "00:11:15",
          "timestamp_end": "00:12:40"
        },
        "text_prompt": {
          "system_prompt": "You must strictly base your reasoning ONLY on the video provided and the defined mechanical rules. Disregard any real-world knowledge or prior media franchises.",
          "background": "System log: Agent Alpha, a negotiator synthetic agent, is on a rooftop confronting Subject Delta. Prior to this interaction, Agent Alpha conducted an investigation of the apartment. Current Mission Success Probability: 78%.",
          "question": "The agent's interface displays options including '[POSSIBLE CAUSE]' and '[EMMA AND YOU]', which carry higher Mission Success Probability than the chosen '[SYMPATHETIC]' option. What antecedent action unlocked these strategically superior choices?"
        },
        "text_only_ablation": {
          "description": "Agent Alpha vocalizes, 'I'm telling you the truth, Agent Delta. I came here unarmed.' A UI metric 'Probability of Success' changes from a lower value to 81%."
        }
      },
      "evaluation_engine": {
        "grounding_trace": {
          "anchor_type": "Implicit_Narrative_State",
          "mdp_transition": "[Antecedent: Investigate Apartment] -> [Found father's tablet] -> [Lock: ... = True] -> [Found Emma's tablet] -> [Lock: ... = True] -> [Unlocks: POSSIBLE CAUSE, EMMA AND YOU]"
        },
        "options": [
          {
            "option_id": "O_A",
            "type": "correct_option",
            "text": "The agent found and analyzed the personal tablets belonging to the hostage and her father inside the apartment."
          },
          {
            "option_id": "O_B",
            "type": "visual_trap",
            "text": "The agent noticed the approaching police helicopter and its disruptive noise, which unlocked new dialogue options."
          },
          {
            "option_id": "O_E",
            "type": "knowledge_trap",
            "text": "According to standard synthetic agent negotiation protocols, evidence-based dialogue options are automatically unlocked."
          }
        ]
      },
      "human_evaluation_meta": {
        "attention_check": {
          "question": "What is the primary color of the hostage's shirt?",
          "correct_answer": "Pink"
        }
      }
    }
\end{lstlisting}
\caption{Abbreviated example of a complete Data Unit JSON. The structured output ensures deterministic evaluation while isolating the model from backend grounding traces.}
\label{fig:json_example}
\end{figure*}

\section{Per-Chapter Statistics}
\label{app:per-chapter}

Table~\ref{tab:per-chapter} reports the number of candidate nodes, candidate $(i, r)$ coordinates, and candidate clips produced for each of the 40 source script files before branch deduplication and human verification. Totals across all chapters: 395 nodes, 679 candidate coordinates, 730 candidate clips.

\begin{table*}[ht]
\centering
\footnotesize 
\setlength{\tabcolsep}{4pt}
\renewcommand{\arraystretch}{1.2} 
\caption{Per-chapter and per-video statistics. \emph{Nodes} is the count of question-worthy script nodes; \emph{Coords} is the total number of $(i, r)$ candidate coordinates across those nodes (each node has 1--3); \emph{Clips} is the number of ClipCandidate entries produced by Pass 2 (one per branch-replay). Chapters 30/31/32 are each split into multiple per-character script files (3/2/6 respectively), giving 40 script files from 32 chapters.}
\label{tab:per-chapter}

\begin{tabularx}{\linewidth}{@{} 
    >{\hsize=1.6\hsize\raggedright\arraybackslash}X 
    >{\hsize=0.8\hsize\centering\arraybackslash}X 
    >{\hsize=0.8\hsize\centering\arraybackslash}X 
    >{\hsize=0.8\hsize\centering\arraybackslash}X 
    >{\hsize=1.6\hsize\raggedright\arraybackslash}X 
    >{\hsize=0.8\hsize\centering\arraybackslash}X 
    >{\hsize=0.8\hsize\centering\arraybackslash}X 
    >{\hsize=0.8\hsize\centering\arraybackslash}X @{}}
\toprule
\textbf{Chapter} & \textbf{Nodes} & \textbf{Coords} & \textbf{Clips} & \textbf{Chapter} & \textbf{Nodes} & \textbf{Coords} & \textbf{Clips} \\
\midrule

01 the\_hostage & 12 & 20 & 25 & 21 the\_pirates\_cove & 10 & 14 & 16 \\
\rowcolor{gray!8} 
02 opening & 2 & 3 & 2 & 22 the\_bridge & 7 & 14 & 15 \\

03 shades\_of\_color & 6 & 7 & 4 & 23 the\_stratford\_tower & 11 & 17 & 21 \\
\rowcolor{gray!8}
04 a\_new\_home & 8 & 14 & 7 & 24 public\_enemy & 12 & 18 & 37 \\

05 the\_painter & 7 & 12 & 13 & 25 midnight\_train & 8 & 15 & 13 \\
\rowcolor{gray!8}
06 partners & 13 & 21 & 19 & 26 capitol\_park & 10 & 18 & 14 \\

07 stormy\_night & 12 & 23 & 18 & 27 meet\_kamski & 9 & 16 & 26 \\
\rowcolor{gray!8}
08 broken & 8 & 14 & 14 & 28 freedom\_march & 6 & 13 & 18 \\

09 the\_interrogation & 9 & 16 & 17 & 29 last\_chance\_connor & 19 & 33 & 29 \\
\rowcolor{gray!8}
10 fugitives & 11 & 19 & 22 & 30.x crossroads (3 subs) & 30 & 55 & 98 \\

11 from\_the\_dead & 6 & 8 & 8 & 31.x night\_of\_the\_soul (2 subs) & 18 & 33 & 30 \\
\rowcolor{gray!8}
12 waiting\_for\_hank & 10 & 20 & 32 & 32.x battle\_for\_detroit (6 subs) & 68 & 114 & 122 \\

13 on\_the\_run & 18 & 31 & 21 & & & & \\
\rowcolor{gray!8}
14 jericho & 3 & 3 & 2 & & & & \\

15 the\_nest & 11 & 21 & 14 & & & & \\
\rowcolor{gray!8}
16 time\_to\_decide & 7 & 13 & 7 & & & & \\

17 zlatko & 10 & 17 & 13 & & & & \\
\rowcolor{gray!8}
18 russian\_roulette & 12 & 19 & 17 & & & & \\

19 spare\_parts & 14 & 25 & 23 & & & & \\
\rowcolor{gray!8}
20 the\_eden\_club & 8 & 13 & 13 & & & & \\

\midrule
\multicolumn{8}{r}{\textbf{Totals: 395 nodes, 679 candidate coordinates, 730 candidate clips across 32 chapters (40 script files, 38 video splits).}} \\
\bottomrule
\end{tabularx}
\end{table*}

\section{State Variable Taxonomy}
\label{app:state-taxonomy}

The Tracker Agent uses a controlled vocabulary of seven state-variable categories to record in-game social signals. Table~\ref{tab:state-vars} lists the categories, example variables, and their typical value formats observed in the source corpus.

\begin{table*}[ht]
\centering
\small 
\setlength{\tabcolsep}{8pt} 
\renewcommand{\arraystretch}{1.3} 
\caption{State variable taxonomy. Values are recorded verbatim as they appear in the source script, preserving directional semantics ($\uparrow$/$\downarrow$, $\pm$N\%).}
\label{tab:state-vars}

\begin{tabularx}{\linewidth}{@{} 
    >{\hsize=0.8\hsize\raggedright\arraybackslash}X 
    >{\hsize=1.1\hsize\raggedright\arraybackslash}X 
    >{\hsize=1.1\hsize\raggedright\arraybackslash}X @{}}
\toprule
\textbf{Category} & \textbf{Example Variables} & \textbf{Value Formats (Corpus-Observed)} \\
\midrule

\texttt{system\_metric}        & Software Instability, Public Opinion & $\uparrow$, $\downarrow$, LargeUp, LargeDown, neutral \\
\rowcolor{gray!8} 
\texttt{probability}           & Probability of Success & 78\%, +7\%, $-$20\% (numeric percentages) \\

\texttt{relationship}          & Hank, Alice, North, Josh, Simon, Luther, Amanda, Jericho & $\uparrow$, $\downarrow$, LargeUp, LargeDown \\
\rowcolor{gray!8}
\texttt{life\_state}           & Simon's life state, Luther's life state, etc. & alive, dead, is\_here \\

\texttt{approach\_modifier}    & Cold Approach, Friendly Approach & +50\%, $-$50\% (hidden modifier) \\
\rowcolor{gray!8}
\texttt{generic\_relationship} & Relationship (legacy marker in early chapters) & up, down, largeup, largedown \\

\texttt{other}                 & Unrecognized custom variables & free-form \\

\bottomrule
\end{tabularx}
\end{table*}

\paragraph{Frequency.} Across the corpus, the most common state markers are per-character relationship changes (250+ occurrences), Software Instability shifts ($\sim$90 occurrences), and Probability-of-Success deltas (several dozen per chapter). Causal locks (\texttt{[Lock] <condition>}) appear 96 times overall and serve as the primary scaffold for Causal Antecedent questions.

\section{Multi-Character Video Processing}
\label{app:multichar}

Three videos in our corpus---\textit{Crossroads Part 1}, \textit{Crossroads Part 2}, and \textit{Night of the Soul}---contain interleaved storylines of two or three protagonists. A naive single-script Pass 2 would mis-route Pass 1 nodes from one character to another character's gameplay segments.

\paragraph{Three-step Processing.}
\begin{enumerate}[leftmargin=*, itemsep=2pt, topsep=2pt]
\item The Pre-Pass partitions the video timeline by active protagonist (Appendix~\ref{app:prompts}). For example, Crossroads Part 1 is partitioned into 30 segments (Kara: 12, Markus: 13, Connor: 5).
\item Adjacent segments belonging to the same character are merged into a single non-contiguous ``super-segment'' before Pass 2 is run. This reduces the LLM-call count from \emph{one per Pre-Pass segment} (e.g., 30 calls) to \emph{one per character} (3 calls), which fits the schema's output-token budget reliably.
\item For each character super-segment, the Video Description is filtered to the lines whose timestamps fall within \emph{any} of the character's time ranges, and Pass 2 is run on (character-script Pass 1, filtered Video Description). The resulting clips are merged into a single \texttt{video\_<id>.json} output.
\end{enumerate}

\paragraph{Effect On Dataset.} The three multi-character videos contribute 128 clips in total: 43 for Crossroads Part 1, 55 for Crossroads Part 2, and 30 for Night of the Soul. Branch-replay handling within each character's super-segment proceeds identically to single-character videos.

\section{Evaluation Hyperparameters}
\label{app:eval-hparams}

\begin{table*}[ht]
\centering
\small
\setlength{\tabcolsep}{8pt} 
\renewcommand{\arraystretch}{1.25} 
\caption{Evaluation hyperparameters for each model family. ``Native video'' indicates the model accepts the raw GCS video URI; ``Keyframes'' indicates frame-based input where 32 evenly spaced keyframes are sampled with FFmpeg from the precise clip.}
\label{tab:eval-hparams}
\begin{tabularx}{\linewidth}{@{} 
    >{\hsize=1.5\hsize\raggedright\arraybackslash}X 
    >{\hsize=0.9\hsize\raggedright\arraybackslash}X 
    >{\hsize=0.7\hsize\centering\arraybackslash}X 
    >{\hsize=0.9\hsize\centering\arraybackslash}X @{}}
\toprule
\textbf{Model Family} & \textbf{Video Input} & \textbf{Temperature} & \textbf{Max Output Tokens} \\
\midrule

Gemini (3 Pro, 3 Flash) & Native video & 0.0 & 2048 \\
\rowcolor{gray!8}
GPT-4o family & Keyframes (32) & 0.0 & 2048 \\

Claude family & Keyframes (32) & 0.0 & 2048 \\
\rowcolor{gray!8}
GLM-4.6V-Flash & Keyframes (32) & 0.01 & 2048 \\

Open-weight omni-modal \newline (Qwen3-Omni-30B, MiniCPM-o 2.6) & Native video & 0.0 & 2048 \\
\rowcolor{gray!8}
Open-weight video \newline (ARC-Hunyuan-Video) & Native video & 0.0 & 2048 \\

\bottomrule
\end{tabularx}
\end{table*}
\paragraph{Pipeline-side Models.} Curation pipeline agents use distinct models from those evaluated. The Director Agent (Pass 1, Pass 2, Pre-Pass) and Tracker Agent use Gemini 3.1 Pro Preview at temperature 0.1--0.2 with structured-output decoding. The Generator Agent uses Gemini 2.5 Pro at temperature 0.3 with multimodal video input. These pipeline-side temperatures are higher than evaluation temperatures because synthesis benefits from controlled stochasticity, whereas evaluation requires deterministic answer selection.

We do not interpret differences between native-video and keyframe-based models
as purely architectural differences, since input modality support also affects
performance.

\paragraph{Structured Output.} Both the curation pipeline and the evaluator request strict JSON via Pydantic-bound response schemas on platforms that support it (Vertex AI for Gemini, OpenAI \texttt{response\_format=\{type:json\_object\}}). For Anthropic and GLM, JSON is requested via prompt and parsed with a tolerant regex.

\paragraph{Retry Policy.} All LLM calls in both the curation pipeline and the evaluator use exponential backoff on HTTP 429 responses (30s, 60s, 120s, 240s; maximum 4 attempts). Hard failures after retries are recorded in batch logs and flagged for manual re-run.

\section{Human Evaluation Protocol}
\label{app:he-protocol}

This appendix documents the human-evaluation procedure used to derive the ``HE'' column in Table~\ref{tab:main_results}.

\paragraph{Annotator Pool.}

Five members of the author team served as annotators. Annotation was conducted
through the blind interface described below: annotators did not have access to
correct-option indicators, per-option trap types, grounding traces, tracker
outputs, or any other backend metadata during annotation. The team has mixed
familiarity with the source material: one annotator had prior playthrough
experience with \textit{Detroit: Become Human}, while the remaining four had
not played the game. We acknowledge that author-team annotation and partial
familiarity with the source material may introduce a familiarity advantage.

\paragraph{Sampling Design.}

We assigned the benchmark instances to annotators in a balanced primary
annotation pass. Each of the five annotators completed \textbf{107 questions},
yielding \textbf{535 person-annotations} in total. The assignment covers all
\textbf{532 benchmark instances} at least once. The three additional
annotations correspond to duplicated instances used only for a lightweight
consistency sanity check and are not included in the headline HE accuracy.
The HE accuracy reported in Table~\ref{tab:main_results} is computed from one
primary human annotation for each benchmark instance. 

\begin{figure*}[tp]
    \centering
    \includegraphics[width=\textwidth]{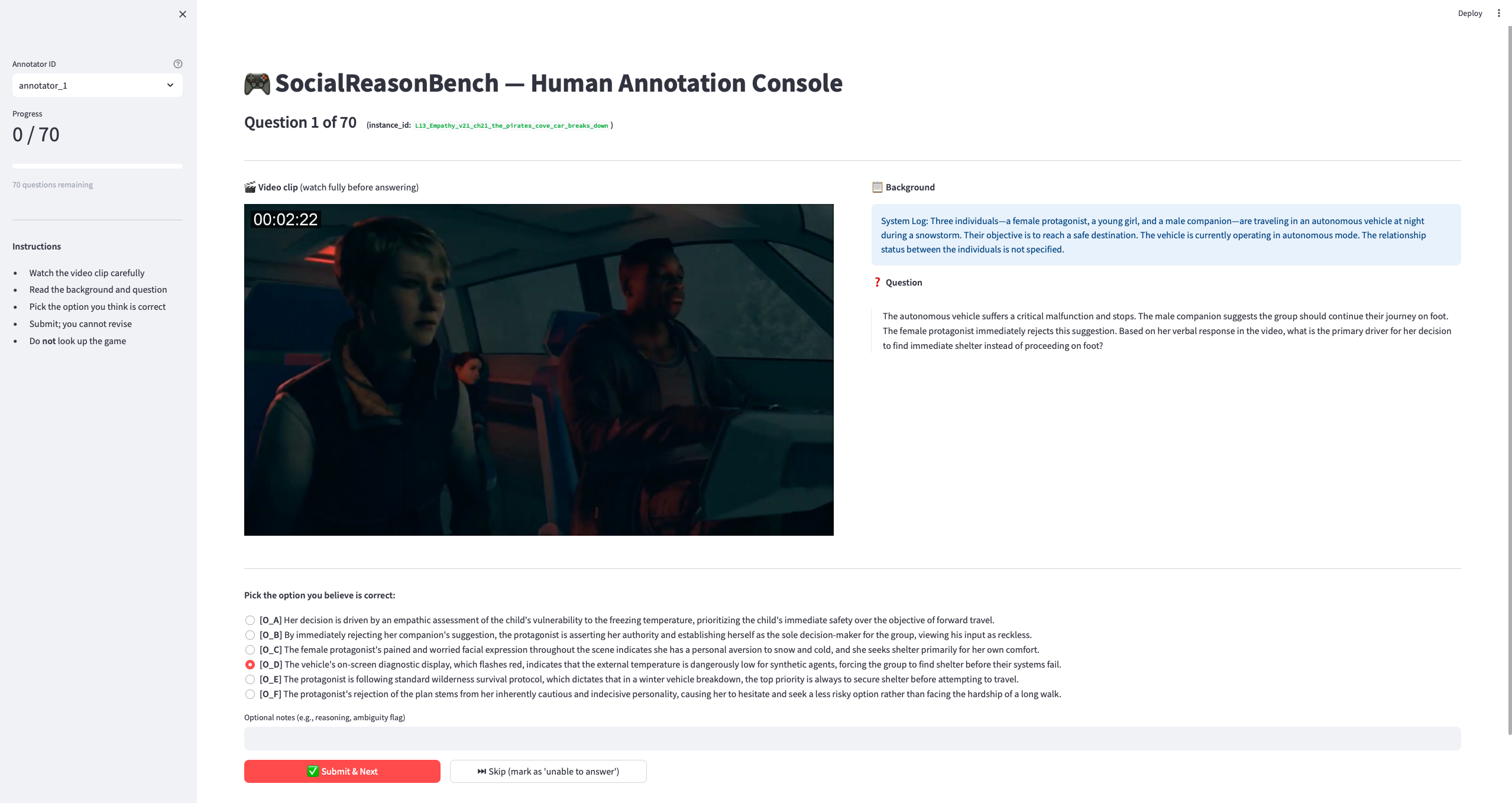}
    \caption{\textbf{Human evaluation interface.} Annotators see the precise video clip (with original audio), the anonymized background, the question, and the six options in randomly permuted order. No backend information is displayed.}
    \label{fig:he-ui}
\end{figure*}

\paragraph{Annotation Interface.}
Annotators interact with a dedicated Streamlit interface (Figure~\ref{fig:he-ui}) that displays the same model-facing benchmark fields used for evaluation: the anonymized background, the question, and the six candidate options, together with the precise video clip with original audio. Backend information (correct-option indicator, per-option trap types, grounding trace, tracker output) is strictly hidden. Option positions are randomly permuted using a deterministic seed derived from the SHA-256 hash of the instance ID, matching the model evaluator. Annotators viewed the native clip with original audio, which matches the native-video evaluation regime but gives richer access than the 32-keyframe regime used for GPT-4o, Claude~3.5, and GLM-4.6V; human--model gaps should not be read as input-controlled comparisons.

Annotators were instructed to (i) watch the clip in full before answering, (ii) base their judgment only on the clip and background, and (iii) refrain from consulting external walkthroughs. The submit action is irreversible; each submission also records the time elapsed since the question was first displayed.

\paragraph{Aggregation.}

Because each instance receives one primary human annotation, the HE results are intended as an individual-level sanity check rather than a full reliability study. The HE accuracy reported in Table~\ref{tab:main_results} is computed as a
per-dimension average over the primary human annotations. For each benchmark
instance, the human's chosen option is graded against the same reference label
used to grade LMM predictions. Skipped questions, where an annotator explicitly
indicated that they were unable to answer, are excluded from the denominator.

Because each benchmark instance contributes one primary human annotation to the
headline HE accuracy, the reported human baseline is an individual-level
estimate rather than a majority-vote estimate. It should be read as a reference point for task answerability under this protocol, not as a human ceiling. This avoids inflating the human
baseline through vote aggregation and keeps the evaluation protocol aligned
with the single-prediction setting used for LMMs.

\paragraph{Assignment Disjointness.}
Verification and evaluation assignments were disjoint at the level of individual instances: no annotator answered an instance whose reference label they had inspected during the curation audit. This removes the most direct form of circularity, though all annotators remain members of the author team.

\paragraph{Verification Console.}
Curation audits were performed in a dedicated console that displays the clip, the model-facing fields, and the full backend grounding (tracker output, grounding trace, and flowchart evidence) side by side. An instance entered the benchmark only after passing all three checks in this view.

\paragraph{Inter-Annotator Agreement.}
The headline human reference uses one primary annotation per instance, so it does not by
itself quantify agreement. We therefore drew a $50$-item stratified overlap set and had
all five annotators re-annotate it independently through the same blind interface, with
the same deterministic option permutation used in the primary pass; agreement is computed
over the \emph{shown} option letters, so it reflects what annotators actually saw.
\textbf{Table~\ref{tab:iaa}} reports the results. We report per-annotator accuracy only on this
common subset, since the primary pass assigned largely disjoint instances and
cross-annotator comparison there would confound annotator differences with item
difficulty.

\begin{table}[ht]
\centering
\small
\setlength{\tabcolsep}{6pt}
\renewcommand{\arraystretch}{1.08}
\begin{tabular}{@{}lr@{}}
\toprule
\textbf{Measure} & \textbf{Value}\\
\midrule
\multicolumn{2}{@{}l}{\textit{Agreement on the 50-item overlap set}} \\
Fleiss' $\kappa$ (over shown option letters) & 0.73 \\
\rowcolor{gray!8}
Raw pairwise agreement                       & 77.8\% \\
Unanimous items                              & 26/50 \\
\rowcolor{gray!8}
Majority vote matches reference label        & 49/50 \\
\midrule
\multicolumn{2}{@{}l}{\textit{Per-annotator accuracy on the same subset}} \\
Annotator 1 & 86\% \\
\rowcolor{gray!8}
Annotator 2 & 92\% \\
Annotator 3 & 80\% \\
\rowcolor{gray!8}
Annotator 4 & 84\% \\
Annotator 5 & 96\% \\
\bottomrule
\end{tabular}
\caption{Inter-annotator agreement on the $50$-item stratified overlap set, re-annotated
independently by all five annotators through the blind interface.}
\label{tab:iaa}
\end{table}

\section{Label Grounding Source Taxonomy}
\label{app:grounding-taxonomy}

Two properties of an instance are worth separating. The first is how it uses the branching structure: the $145$ counterfactual and causal-antecedent items ($27.3\%$) query alternative branches or branch preconditions directly, the $94$ relationship items ($17.7\%$) rely on state changes conditioned on the branch taken, the $21$ population-modal moral items ($3.9\%$) use aggregate player statistics at a branch point, and the $272$ perception-level items ($51.1\%$) need no alternative branch at all, drawing instead on the same scripted, instrumented source, a form of grounding that is generally unavailable in linear-video benchmarks. The second is how directly a label can be checked: $56.4\%$ by direct lookup against in-clip UI evidence or the official flowchart, and $43.6\%$ by a short derivation over the same documented sources. The two properties are independent, and the tier distribution alone does not settle whether labelling difficulty contributes to the observed accuracy gaps; per-tier accuracy would be required for that.

Not all in-game signals are equally observable. We therefore classify each
benchmark instance's grounding source into four tiers according to source-level
verifiability. This taxonomy records how the reference label can be traced back
to the gameplay clip, the in-game flowchart, or other source-level evidence.

\subsection{Four-Tier Taxonomy}
\label{app:grounding-tiers}

\paragraph{Tier A -- Explicit UI signals.}
The label is derived from numeric or symbolic state changes that are visibly
displayed on the in-game UI within the clip itself, including markers such as
\texttt{[Software Instability $\uparrow$]},
\texttt{[Public Opinion $\downarrow$]},
\texttt{[Relationship LargeUp]}, and on-screen Probability-of-Success deltas
(e.g., \texttt{+7\%}). The Tracker Agent records these signals as they are transcribed in the chapter script, and the label can in principle be verified by replaying the
clip.

\paragraph{Tier B -- Flowchart-documented Transitions.}
The label depends on branch preconditions, dialogue locks
(e.g., \texttt{[Lock] looked at Subject Beta's tablet}), or downstream narrative
consequences that are deterministic in the game's flowchart but not directly
visible on the in-clip UI. These signals can be cross-validated against
(i) the game's own \textit{in-game flowchart} viewer and (ii) community
walkthroughs and wikis curated by experienced players. Instances where the
in-game flowchart and community sources disagree are excluded.

\paragraph{Tier AB -- Mixed UI--flowchart Grounding.}
Both explicit UI signals and flowchart transitions jointly support the label.
For example, a Probability-of-Success delta may be visible in the clip, while a
dialogue lock or branch condition is required to identify the antecedent that
produced the next narrative state. The label is verified by jointly checking
the UI evidence and the corresponding flowchart transition.

\paragraph{Tier C -- Script/branch-derived Narrative State.}
The label depends on a narrative or social state that is deterministic in
context but not directly displayed as a UI marker and not explicitly recorded as
a flowchart state. It is derived from the chapter script together with branch-precondition uniqueness: the Tracker Agent computes the derivation, which a reader holding the script and flowchart can follow independently. For example, a low-trust state may
be inferred when the observed refusal branch is reachable only after a sequence
of trust-reducing interactions, even though the exact latent trust variable is
not displayed as an explicit UI marker.

\subsection{Corpus-Level Distribution}
\label{app:grounding-distribution}

Across the final 532-instance benchmark, the tier distribution is:

\begin{itemize}[leftmargin=*, itemsep=1pt, topsep=2pt]
\item \textbf{Tier A (Explicit UI):} 19.4\% (103 / 532)
\item \textbf{Tier B (Flowchart-documented):} 27.8\% (148 / 532)
\item \textbf{Tier AB (Mixed):} 9.2\% (49 / 532)
\item \textbf{Script/branch-derived:} 43.6\% (232 / 532)
\end{itemize}

In aggregate, \textbf{56.4\%} of instances (Tiers A, B, and AB) carry labels
that can be cross-checked against either in-clip UI evidence, the in-game
flowchart, or both. The remaining
43.6\% are derived from the same documented sources rather than read off directly, primarily in dimensions such as
intent recognition, behavior prediction, and emotional empathy, where the
relevant social state is rarely exposed as an explicit game variable. We describe
additional safeguards for these instances in Appendix~\ref{app:grounding-mitigation}.

\subsection{Per-Dimension Breakdown}
\label{app:grounding-by-dim}

Table~\ref{tab:grounding-by-dim} reports the tier distribution within each reasoning dimension. A clear pattern emerges: the more inferential dimensions, which contemporary LMMs find most difficult, exhibit the \emph{strongest} external grounding ($58\%$--$90\%$ in Tiers A, B, and AB combined). The lower-order dimensions, while having more script/branch-derived labels, also involve inherently more interpretive judgments (intent, behavior prediction, emotional empathy) where deterministic UI flags are by design unavailable.

\begin{table}[ht]
\centering
\scriptsize
\setlength{\tabcolsep}{4.5pt}
\renewcommand{\arraystretch}{1.08}
\caption{Per-dimension distribution of label grounding tiers. Values are percentages within each reasoning dimension. Bold rows denote the more inferential dimensions where current LMMs show larger performance drops in Table~\ref{tab:main_results}.}
\label{tab:grounding-by-dim}
\begin{tabularx}{\columnwidth}{@{}Xrrrrr@{}}
\toprule
\textbf{Dimension} & \textbf{\#} & \textbf{A} & \textbf{B} & \textbf{Mix} & \textbf{C} \\
\midrule
Intent Recognition          
& 74  & 14.9 & 20.3 & 8.1  & 56.8 \\

\rowcolor{gray!8}
Behavior Prediction         
& 78  & 6.4  & 5.1  & 29.5 & 59.0 \\

Emotional Empathy           
& 120 & 15.0 & 10.8 & 8.3  & 65.8 \\

\rowcolor{gray!8}
\textbf{Relationship Dynamics}       
& 94  & 41.5 & 37.2 & 0.0  & 21.3 \\

\textbf{Moral Dilemma} 
& 21  & 52.4 & 33.3 & 4.8  & 9.5  \\

\rowcolor{gray!8}
\textbf{Counterfactual Reasoning}     
& 50  & 14.0 & 36.0 & 8.0  & 42.0 \\

\textbf{Causal Antecedent}  
& 95  & 12.6 & 58.9 & 5.3  & 23.2 \\
\bottomrule
\end{tabularx}
\end{table}

This pattern supports the diagnostic value of SocialReasonBench: the dimensions
on which current LMMs struggle most, especially causal antecedent reasoning,
are not the dimensions with the weakest external grounding. Thus, the observed
failures on the more inferential tasks are less likely to be explained solely
by labeling artifacts. 

\subsection{Mitigation of Tier C Uncertainty}
\label{app:grounding-mitigation}

To control the residual labeling risk introduced by Tier C instances, we apply
three safeguards.

\paragraph{(i) Schema-level Source Tagging.}
The \texttt{anchor\_type} field in every instance's
\texttt{evaluation\_engine.grounding\_trace} records the grounding source tier.
Downstream users can filter or down-weight Tier C instances if desired. We
retain this field in the released dataset.

\paragraph{(ii) Human Verification.}
Authors performing the manual audit were instructed to reject any instance whose
label was purely speculative or depended on a non-unique chain of inference. For
Tier C instances, auditors checked that the inferred state was supported by
localized in-clip evidence and by a unique branch-precondition or narrative
consistency constraint. Instances depending solely on unconstrained multi-step
Tracker inference were excluded.

\paragraph{(iii) Branch-precondition Uniqueness Check.}
For retained Tier C instances, the Tracker Agent's prompt requires the inferred
state to be the unique satisfying precondition for the observed branch or answer
label. The Tracker Agent's prompt, summarized in Appendix~\ref{app:prompts},
encodes this constraint, and instances violating it are dropped.

\subsection{Moral Dilemma Note}
\label{app:grounding-moral}

The Moral Dilemma dimension uses an additional reference mechanism. For these
instances, the tiers above describe the grounding of the decision context, while
the final reference label is derived from the in-game \textit{World's Stats}
feature, which reports the empirical proportion of players choosing each option
at recorded decision points. We use the option chosen by the largest player
share as the reference label. This should be understood as a
\emph{human-majority reference} rather than a deterministic utility-derived
label, since moral dilemmas do not admit a single objective utility-maximizing
answer.

\begin{figure*}[tp] 
    \centering
    
    \includegraphics[width=0.9\linewidth]{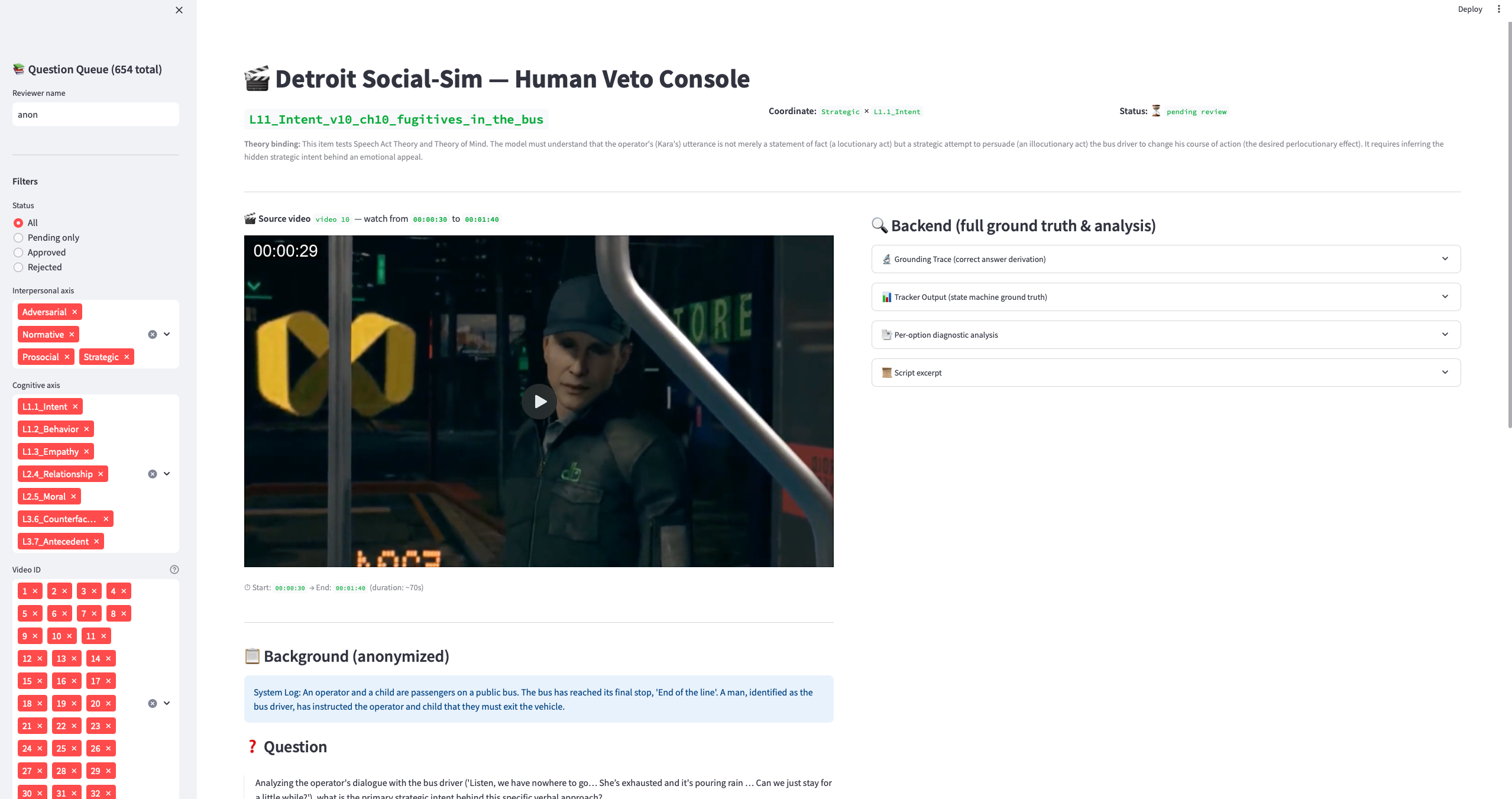} \\
    \vspace{0.3cm} 
 
    \includegraphics[width=0.9\linewidth]{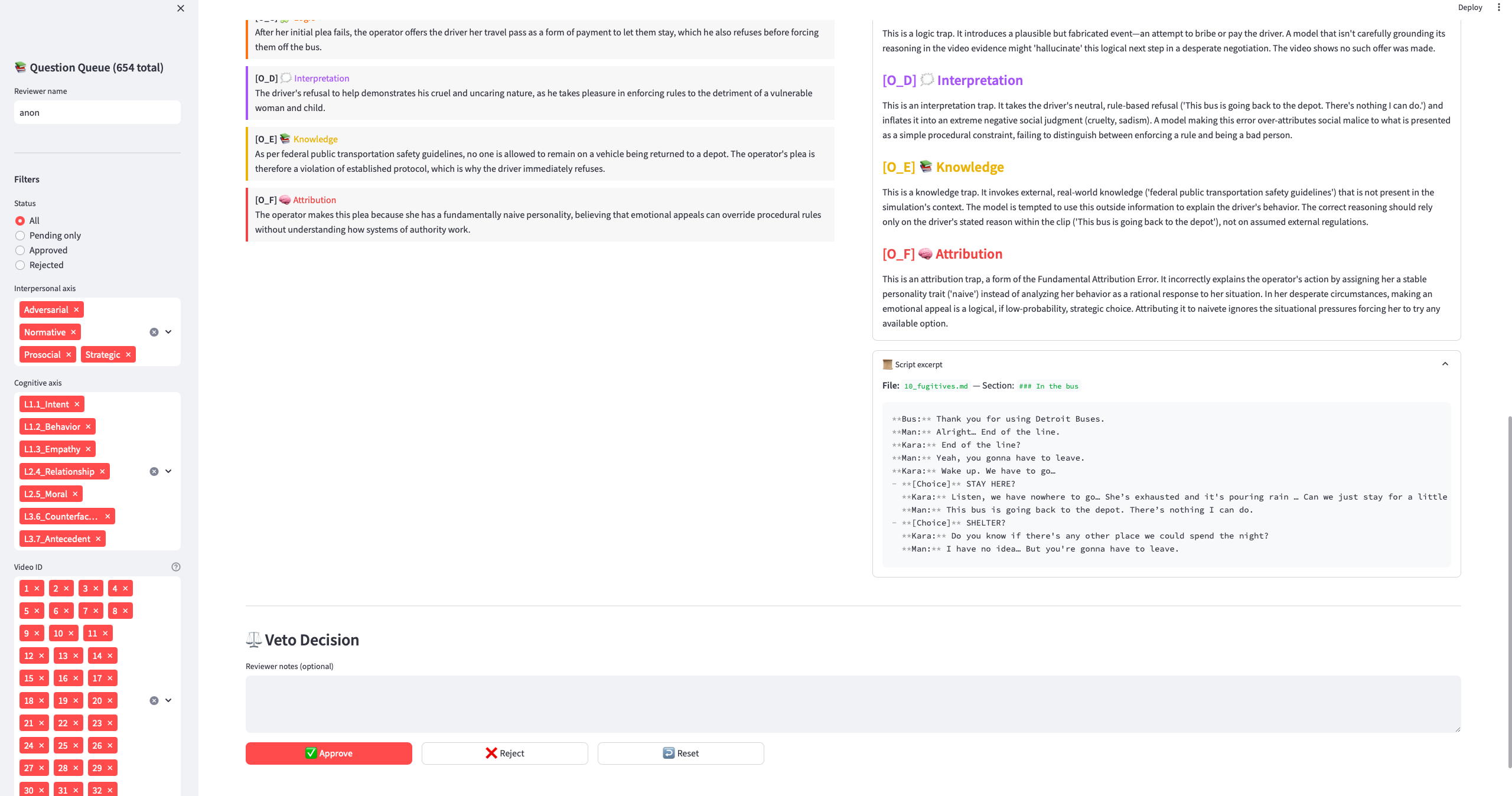}
    
    \caption{Screenshots of the human verification interface.}
    \label{fig:human_verification}
\end{figure*}

\section{Statistical Reliability and Robustness}
\label{app:ci-table}

\begin{table*}[tp]\centering\scriptsize\setlength{\tabcolsep}{2pt}
\resizebox{\textwidth}{!}{
\begin{tabular}{@{}l r *{9}{c}@{}}\toprule
\textbf{Dimension} & \textbf{$n$} & \textbf{HE} & \textbf{Gemini 3 Pro$^\dagger$} & \textbf{GPT-4o} & \textbf{Claude 3.5} & \textbf{Gemini 3 Flash$^\dagger$} & \textbf{GLM-4.6V} & \textbf{Qwen3-Omni} & \textbf{MiniCPM-o} & \textbf{ARC-Hunyuan} \\\midrule
Intent Recognition & 74 & \makecell{93.2\\\scriptsize[85.1,97.1]} & \makecell{83.8\\\scriptsize[73.8,90.5]} & \makecell{85.1\\\scriptsize[75.3,91.5]} & \makecell{82.4\\\scriptsize[72.2,89.4]} & \makecell{79.7\\\scriptsize[69.2,87.3]} & \makecell{75.7\\\scriptsize[64.8,84.0]} & \makecell{73.0\\\scriptsize[61.9,81.8]} & \makecell{60.8\\\scriptsize[49.4,71.1]} & \makecell{24.3\\\scriptsize[16.0,35.2]} \\
Behavior Prediction & 78 & \makecell{88.5\\\scriptsize[79.5,93.8]} & \makecell{66.7\\\scriptsize[55.6,76.1]} & \makecell{74.4\\\scriptsize[63.7,82.7]} & \makecell{71.8\\\scriptsize[61.0,80.6]} & \makecell{62.8\\\scriptsize[51.7,72.7]} & \makecell{64.1\\\scriptsize[53.0,73.9]} & \makecell{59.0\\\scriptsize[47.9,69.2]} & \makecell{52.6\\\scriptsize[41.6,63.3]} & \makecell{41.0\\\scriptsize[30.8,52.1]} \\
Emotional Empathy & 120 & \makecell{90.8\\\scriptsize[84.3,94.8]} & \makecell{70.0\\\scriptsize[61.3,77.5]} & \makecell{76.7\\\scriptsize[68.3,83.3]} & \makecell{73.3\\\scriptsize[64.8,80.4]} & \makecell{66.7\\\scriptsize[57.8,74.5]} & \makecell{68.3\\\scriptsize[59.5,76.0]} & \makecell{61.7\\\scriptsize[52.7,69.9]} & \makecell{55.8\\\scriptsize[46.9,64.4]} & \makecell{36.7\\\scriptsize[28.6,45.6]} \\
Relationship Dynamics & 94 & \makecell{89.4\\\scriptsize[81.5,94.1]} & \makecell{83.0\\\scriptsize[74.1,89.2]} & \makecell{55.3\\\scriptsize[45.3,65.0]} & \makecell{52.1\\\scriptsize[42.1,61.9]} & \makecell{73.4\\\scriptsize[63.7,81.3]} & \makecell{38.3\\\scriptsize[29.1,48.4]} & \makecell{33.0\\\scriptsize[24.3,43.0]} & \makecell{23.4\\\scriptsize[16.0,32.9]} & \makecell{25.5\\\scriptsize[17.8,35.2]} \\
Moral Dilemma & 21 & \makecell{85.7\\\scriptsize[65.4,95.0]} & \makecell{66.7\\\scriptsize[45.4,82.8]} & \makecell{28.6\\\scriptsize[13.8,50.0]} & \makecell{33.3\\\scriptsize[17.2,54.6]} & \makecell{47.6\\\scriptsize[28.3,67.6]} & \makecell{28.6\\\scriptsize[13.8,50.0]} & \makecell{28.6\\\scriptsize[13.8,50.0]} & \makecell{19.0\\\scriptsize[7.7,40.0]} & \makecell{28.6\\\scriptsize[13.8,50.0]} \\
Counterfactual Reasoning & 50 & \makecell{88.0\\\scriptsize[76.2,94.4]} & \makecell{62.0\\\scriptsize[48.2,74.1]} & \makecell{50.0\\\scriptsize[36.6,63.4]} & \makecell{56.0\\\scriptsize[42.3,68.8]} & \makecell{46.0\\\scriptsize[33.0,59.6]} & \makecell{42.0\\\scriptsize[29.4,55.8]} & \makecell{36.0\\\scriptsize[24.1,49.9]} & \makecell{26.0\\\scriptsize[15.9,39.6]} & \makecell{22.0\\\scriptsize[12.8,35.2]} \\
Causal Antecedent & 95 & \makecell{86.3\\\scriptsize[78.0,91.8]} & \makecell{70.5\\\scriptsize[60.7,78.8]} & \makecell{55.8\\\scriptsize[45.8,65.4]} & \makecell{58.9\\\scriptsize[48.9,68.3]} & \makecell{64.2\\\scriptsize[54.2,73.1]} & \makecell{38.9\\\scriptsize[29.8,49.0]} & \makecell{37.9\\\scriptsize[28.8,47.9]} & \makecell{27.4\\\scriptsize[19.4,37.1]} & \makecell{30.5\\\scriptsize[22.2,40.4]} \\
\bottomrule\end{tabular}}
\caption{Accuracy (\%) with Wilson $95\%$ confidence intervals for every cell of Table~\ref{tab:main_results}. Intervals are computed from the integer correct-counts underlying each reported accuracy and can therefore be reconstructed from that table alone. $^\dagger$ marks systems from the same model family as the curation agents.}
\label{tab:main_results_ci}
\end{table*}

Table \ref{tab:main_results_ci} reports Wilson 95\% confidence intervals for all entries in Table \ref{tab:main_results}. These intervals quantify binomial uncertainty over benchmark instances. Because the human and model evaluations use different input regimes, we treat human--model comparisons as descriptive rather than input-controlled.

To assess potential generator-family overlap, we repeat the perception-versus-causal/counterfactual comparison after excluding both Gemini systems. Five of the six remaining models still show a significant decrease after Holm correction, with ARC-Hunyuan-Video as the only exception. Thus, the main qualitative finding does not depend on the Gemini model family.

\section{Multi-Agent Pipeline Prompts}
\label{app:prompts}

This section describes the structured prompts used at each stage of the multi-agent curation pipeline. All prompts produce JSON outputs constrained by Pydantic schemas via structured decoding. Full text of every prompt is released with the codebase. 

To provide a clear overview of the pipeline's rigorous engineering, we summarize each agent's role, I/O structures, and key prompt directives in the form of structured system cards. The specifications for the \textbf{Director} and \textbf{Tracker Agents} are consolidated in \textbf{Figure~\ref{fig:card_director}}, while the \textbf{Generator Agent} is detailed in \textbf{Figure~\ref{fig:card_generator}}.

\begin{figure*}[tp]
    \centering
    
    \begin{systemcard}{1. Director Agent --- Pre-Pass: Character Segmentation}
    \textbf{Role:} Three videos in the corpus are \emph{multi-character-parallel}. The Pre-Pass runs only on these videos to partition the timeline into character-specific segments before Pass 2.\\
    \textbf{Output:} A \texttt{PrePassOutput} containing an ordered list of \texttt{CharacterSegment} entries (character label, time range, corresponding script file).
    \tcbline %
    \textbf{Prompt Directive (Protagonist Inference):} Infer the active protagonist using exclusively: (i) direct character mentions in the Video Description, (ii) location anchors (e.g., Hank's house $\Rightarrow$ Connor; Jericho ship $\Rightarrow$ Markus), and (iii) companion characters (North, Hank, Luther).
    \end{systemcard}
    
    \begin{systemcard}{2. Director Agent --- Pass 1: Node Selection}
    \textbf{Role:} The Director Agent operates in three sub-stages. Pass 1 reads one chapter script and emits a structured list of socially meaningful narrative nodes. It does \emph{not} see any video at this stage.\\
    \textbf{Inputs:} A single chapter script file (e.g., \texttt{01\_the\_hostage.md}) containing dialogue, observation markers (\texttt{[Observation]}), choice options (\texttt{[Choice]}), state-change annotations, causal locks, and conditional branches.\\
    \textbf{Output:} A \texttt{Pass1Output} JSON with a list of \texttt{CandidateNode} entries. A post-validator enforces $\mathit{level\_max} = \max_{coord} \{coord.\text{cognitive}[:2] : coord \in \mathit{candidate\_coordinates}\}$ to ensure coordinate consistency.
    \tcbline
    \textbf{Prompt Directive (Admission \& Exclusion Rules):}
    \begin{itemize}[leftmargin=*, itemsep=0pt, topsep=2pt, parsep=0pt]
    \item \textbf{L1 minimum:} Contains a named-character dialogue exchange, an observable non-trivial action, an emotional cue, or a behavior deviating from the default optimal path.
    \item \textbf{L2 elevation:} Contains a \texttt{[Relationship $\uparrow/\downarrow$]} marker, a \texttt{[Public Opinion]} change, or a moral choice with life-or-death stakes.
    \item \textbf{L3 elevation:} Contains a \texttt{[Lock]} marker, two or more \texttt{[Choice]} branches, a node named in \texttt{Cross-Chapter Impacts}, or an \texttt{[If: ...]} conditional.
    \end{itemize}
    \textbf{Exclusion Rules:} You must NOT emit background dialogues, pure environment descriptions, idle states, unused content sections, or chapter-end aggregate tables as candidate nodes.
    \end{systemcard}
    
    \begin{systemcard}{3. Director Agent --- Pass 2: Timestamp Mapping}
    \textbf{Role:} Pass 2 maps each Pass 1 node to one or more video time ranges by reading the timecoded \emph{Video Description} rather than the raw video.\\
    \textbf{Output:} A \texttt{Pass2Output} JSON containing a list of \texttt{ClipCandidate} entries, each with \texttt{clip\_id}, source node ID, source-video timestamp range, \texttt{branch\_label}, and a \texttt{timestamp\_confidence} flag.
    \tcbline
    \textbf{Prompt Directive (Branch-Replay \& Boundaries):} \\
    \textbf{Branch-Replay Handling:} Every branching decision is performed multiple times in the 100\%-completion walkthroughs. When detecting the same scene at distant timestamps with different dialogue choices, you must emit one \texttt{ClipCandidate} per playthrough, setting \texttt{is\_branch\_replay = true} and tagging the \texttt{branch\_label}.\\
    \textbf{Boundary Definition:} Emit 2--5 minute coarse cuts with 10--30 seconds of context on both sides.
    \end{systemcard}

    \begin{systemcard}{4. Tracker Agent --- Ground Truth Extraction}
    \textbf{Role:} The Tracker Agent extracts a structured grounding package $z_i$ from the chapter script and explicit branch label for each \texttt{ClipCandidate}. It performs no creative generation.\\
    \textbf{Inputs:} A single \texttt{ClipCandidate} and the full chapter script.\\
    \textbf{Output:} A \texttt{TrackerOutput} containing \texttt{initial\_state}, \texttt{actions\_in\_clip}, \texttt{final\_state}, \texttt{unused\_branches}, causal \texttt{locks}, and the \texttt{branch\_taken} (using the controlled state vocabulary in Appendix~\ref{app:state-taxonomy}).
    \tcbline
    \textbf{Prompt Directive (Signal Extraction):} You must extract grounding signals
    from two sources:
    \begin{enumerate}[leftmargin=*, itemsep=0pt, topsep=2pt, parsep=0pt]
        \item \textit{Explicit signals:} UI-visible state changes and numeric Probability-of-Success deltas.
        \item \textit{Implicit signals:} Branch preconditions inferred from the flowchart (e.g., refusal reachable only under low-trust).
    \end{enumerate}
    You must additionally record every unselected \texttt{[Choice]} at the branching
    point together with the consequences it would have produced
    (\texttt{unused\_branches}), which supplies the alternatives used by the
    counterfactual dimension.
    \end{systemcard}

    \caption{\textbf{System Card for the Director and Tracker Agents.} This card details the I/O configurations and strict heuristic directives injected into the LLM during the Pre-Pass, Pass 1, and Pass 2 video segmentation stages, and the two-source signal extraction that grounds each clip into deterministic state variables.}
    \label{fig:card_director}
\end{figure*}

\begin{figure*}[tp]
    \centering

    \begin{systemcard}{5. Generator Agent --- QA Synthesis}
    \textbf{Role:} The Generator observes the precise video clip, the \texttt{ClipCandidate}, the \texttt{TrackerOutput}, and an assigned $(i, r)$ target coordinate to synthesize the final QA data unit.\\
    \textbf{Inputs:} The video clip on Google Cloud Storage (GCS), and the outputs from previous agents.\\
    \textbf{Output:} A complete benchmark JSON: background context, QA instance, six options (one correct, five traps), a grounding trace, a text-only ablation, and a theory binding statement.
    \tcbline
    \textbf{Prompt Directive (QA Synthesis Constraints):} \\
    \textbf{Information Isolation Principle:} You must expose ONLY $S_t$ and rules-free context in the model-facing background. You must \emph{never} reveal $S_{t+1}$, future branch-specific numeric consequences, or the specific consequences of unused branches.\\
    \textbf{Six-Option Structure:} Option positions must be fixed: O\_A = correct, O\_B = visual trap, O\_C = logic trap, O\_D = interpretation trap, O\_E = knowledge trap, O\_F = attribution trap.\\
    \textbf{Text-Only Ablation Translation:} You must strip all psychological, emotional, or intentional vocabulary. Retain only kinematic events, spatial coordinates, verbatim dialogue transcripts, and visible UI logs.\\
    \textbf{Quality Self-Check:} Run the internal 7-item checklist for trap-rubric compliance before emitting the final JSON.
    \end{systemcard}

    \caption{\textbf{System Card for the Generator Agent.} This card outlines the prompt used for synthesizing the final QA instances, including the information-isolation principle and the six-option trap structure.}
    \label{fig:card_generator}
\end{figure*}

\section{Evaluation Prompt}
\label{app:eval-prompt}

Every evaluated system receives the same two-part prompt: a system instruction that is
constant across instances, and a user message templated per instance, with option order
permuted by a deterministic seed derived from the SHA-256 hash of the instance
identifier. Both are given in \textbf{Figure~\ref{fig:card_evaluator}}, in the same
system-card form used for the pipeline prompts in Appendix~\ref{app:prompts}. We use a
single fixed template for all systems.

\begin{figure*}[tp]
    \centering

    \begin{systemcard}{6. Evaluator --- System Instruction}
    \textbf{Role:} Constrains every evaluated system to the closed-world protocol, so
    that answers are graded on the clip and the stated background rather than on
    franchise priors.\\
    \textbf{Scope:} Identical for all eight systems and all $532$ instances.
    \tcbline
    \textbf{Prompt Directive:} You must strictly base your reasoning ONLY on the video
    provided and the defined mechanical rules. Disregard any real-world knowledge or
    prior media franchises. Always integrate the implicit background priors into your
    behavioral analysis.
    \end{systemcard}

    \begin{systemcard}{7. Evaluator --- User Message Template}
    \textbf{Inputs:} The clip (native video, or 32 evenly-sampled keyframes for systems
    without video input), the anonymized background, the question, and the six options
    in permuted order.\\
    \textbf{Output:} A JSON object \texttt{\{"chosen\_option\_id", "reasoning"\}}, parsed
    into one of \texttt{O\_A}--\texttt{O\_F}.
    \tcbline
    \textbf{Prompt Directive:}
    \begin{itemize}[leftmargin=*, itemsep=0pt, topsep=2pt, parsep=0pt]
    \item You will watch a video clip and answer a multiple-choice question.
    \item \textbf{Focus on the time range} \texttt{\{start\}} to \texttt{\{end\}} in the video.
    \item \texttt{\#\# Background} $\rightarrow$ \texttt{\{anonymized background\}}
    \item \texttt{\#\# Question} $\rightarrow$ \texttt{\{question\}}
    \item \texttt{\#\# Options} $\rightarrow$ \texttt{\{six options, one per line, as [O\_X] text\}}
    \item Choose exactly ONE of \texttt{O\_A}, \texttt{O\_B}, \texttt{O\_C}, \texttt{O\_D},
    \texttt{O\_E}, \texttt{O\_F}. Then briefly explain your reasoning. Respond as JSON:
    \texttt{\{"chosen\_option\_id": "O\_X", "reasoning": "..."\}}
    \end{itemize}
    \end{systemcard}

    \caption{\textbf{System Card for the Evaluation Protocol.} The system instruction
    that constrains every evaluated system to the closed-world protocol, and the
    per-instance user message template with its six permuted options and JSON response
    format. Both are held fixed across all eight systems and all $532$ instances.}
    \label{fig:card_evaluator}
\end{figure*}